\documentclass[10pt,twocolumn,letterpaper]{article}

\usepackage[pagenumbers]{iccv} 

\definecolor{iccvblue}{rgb}{0.21,0.49,0.74}
\usepackage[pagebackref,breaklinks,colorlinks,allcolors=iccvblue]{hyperref}

\def\paperID{3984} 
\def\confName{ICCV}
\def\confYear{2025}

\title{Can3Tok: Canonical 3D Tokenization and Latent Modeling of \\ Scene-Level 3D Gaussians}

\author{Quankai Gao$^{1}$ \hspace{2mm} Iliyan Georgiev$^{2}$ \hspace{2mm} Tuanfeng Y. Wang$^{2}$ \hspace{2mm} Krishna Kumar Singh$^{2}$ \hspace{2mm} \\ Ulrich Neumann$^{1,\dag}$ \hspace{2mm} Jae Shin Yoon$^{2,\dag}$  \\
    \hspace{0mm}$^1$University of Southern California \hspace{18mm} 
    $^2$Adobe Research\\
}

\begin{document}
\maketitle

\let\thefootnote\relax\footnotetext{$^{\dag}$Equal advising.}

\begin{abstract}
3D generation has made significant progress, however, it still largely remains at the object-level. Feedforward 3D scene-level generation has been rarely explored due to the lack of models capable of scaling-up latent representation learning on 3D scene-level data. Unlike object-level generative models, which are trained on well-labeled 3D data in a bounded canonical space, scene-level generations with 3D scenes represented by 3D Gaussian Splatting (3DGS) are unbounded and exhibit scale inconsistency across different scenes, making unified latent representation learning for generative purposes extremely challenging. In this paper, we introduce Can3Tok, the first 3D scene-level variational autoencoder (VAE) capable of encoding a large number of Gaussian primitives into a low-dimensional latent embedding, which effectively captures both semantic and spatial information of the inputs. Beyond model design, we propose a general pipeline for 3D scene data processing to address scale inconsistency issue. We validate our method on the recent scene-level 3D dataset DL3DV-10K, where we found that only Can3Tok successfully generalizes to novel 3D scenes, while compared methods fail to converge on even a few hundred scene inputs during training and exhibit zero generalization ability during inference. Finally, we demonstrate image-to-3DGS and text-to-3DGS generation as our applications to demonstrate its ability to facilitate downstream generation tasks. Project page: \href{https://github.com/Zerg-Overmind/Can3Tok}{https://github.com/Zerg-Overmind/Can3Tok}
\vspace{-5mm}
\end{abstract}    
\section{Introduction}
\label{sec:intro}

\begin{figure}[t!]
    \centering
    \begin{minipage}[t]{1.0\linewidth} 
        \centering
        \includegraphics[width=\linewidth]{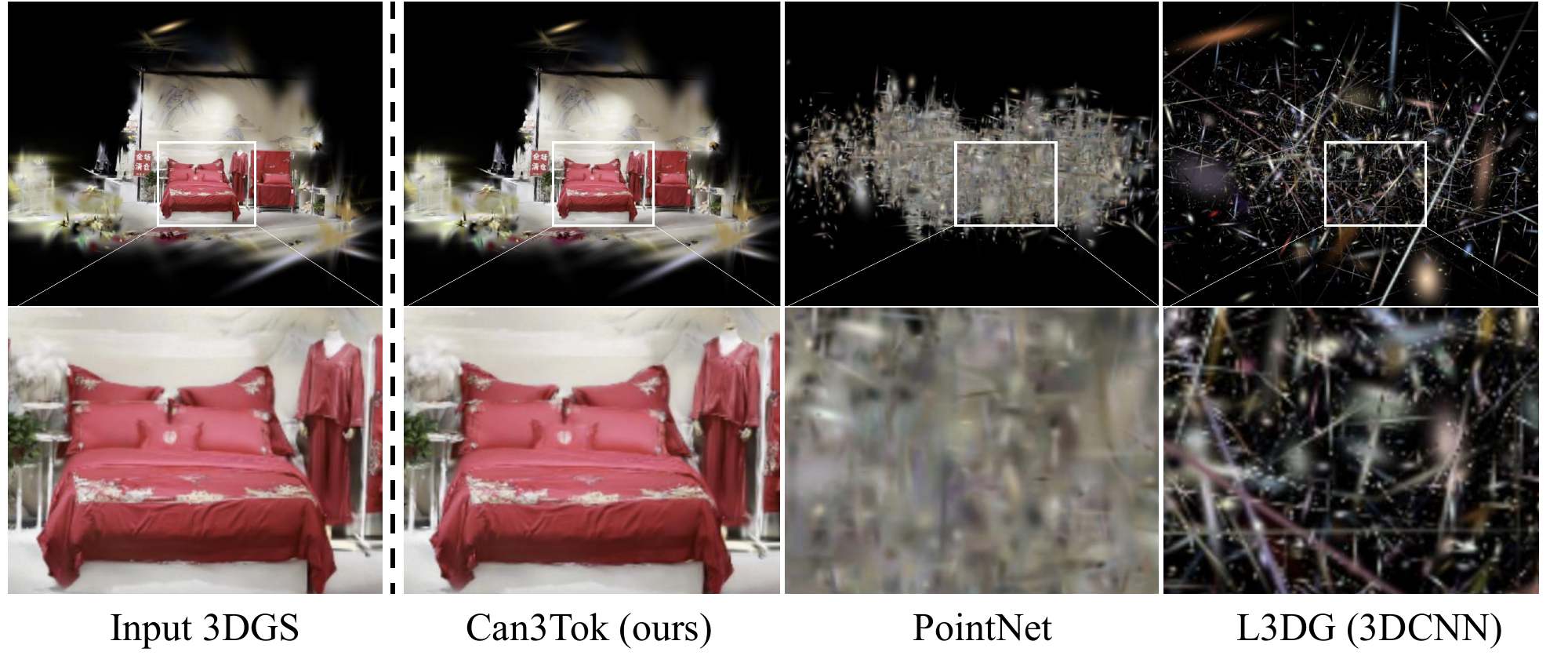} 
    \end{minipage}
    \vspace{-7mm}
    \caption{
        Reconstruction results from the latent space of 3D Gaussian splats (3DGS) of a general scene using our Can3Tok and other 3D-based VAE models~\cite{roessle2024l3dg,qi2017pointnet}. Can3Tok effectively preserves the global shape and local details while existing methods often fail to model the structured latent space from unstructured 3DGS.   
    }
    \vspace{-5mm}
    \label{fig:teaser}
\end{figure}
Realistic 3D Scene-level generation enables immersive AR/VR applications.    While tremendous progress has been made in 3D object-level generation using various 3D representations~\cite{gao2022get3d,nichol2022point,poole2022dreamfusion,tang2023dreamgaussian,hong2023lrm}, with significant improvements ranging from per-scene optimization to feedforward approaches, less research has focused on 3D scene-level generation. Pioneering works such as WonderJourney~\cite{yu2024wonderjourney} and LucidDreamer~\cite{chung2023luciddreamer} generate 3D scene content without training on 3D scene-level data. However, their per-scene optimization is time-consuming, prone to to texture saturation, and lacks 3D consistency due to the using of 2D diffusion models. Our approach explores realistic feedforward 3D scene-level generation by directly training on real 3D data. Conceptually, it aligns with the idea of Stable Diffusion~\cite{rombach2022high}, which enables conditional generation by aligning different modalities (e.g., text and 2D images) within the same latent space for controllable generation. Since the diffusion process and architectures such as UNet or DiT~\cite{peebles2023scalable} are well-known to be grounded, the key bottleneck we identified in feedforward 3D generation is the development of a 3D VAE for learning scene-level latent representations. However, we find this to be non-trivial, as scene-level content is not merely a combination of multiple object-level elements but also includes the background, scene layout, and the relative position, scale, and orientation of objects.

\begin{figure}[t!]
    \centering
    \begin{minipage}[t]{1.0\linewidth} 
        \centering
        \includegraphics[width=\linewidth]{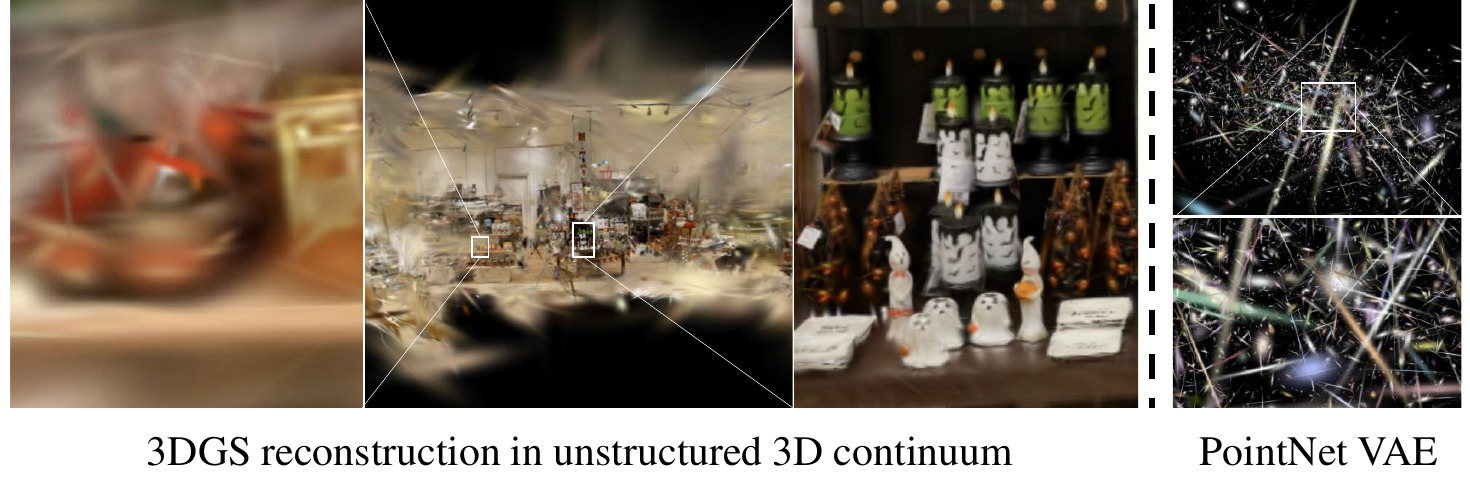} 
    \end{minipage}
    \vspace{-6mm}
    \caption{
        Example of unstructured 3DGS representation of a scene (left) and its decoding results using PointNet-based VAE~\cite{qi2017pointnet,qi2017pointnetplusplus} (right). 3DGS is highly unstructured in continuous and irregular 3D continuum where it often includes many floaters for the region with insufficient multi-view observations.
    }
    \vspace{-3mm}
    \label{fig:motivation}
\end{figure}

In this paper, we explore to learn a new VAE that can project 3D data into a structured latent space. 3D Gaussian Splatting~\cite{kerbl20233d} (3DGS) is an emerging 3D representation that describes a scene as a set of Gaussians with a few parameters such as position, scaling values, and etc. 
One might wonder, is it still possible to build a latent space for many input 3DGS scene representations that are decodable with existing 3D-based VAE models, such as PointNet VAE? Our experiments revealed that existing VAEs cannot reconstruct the input 3DGS as shown in \cref{fig:teaser}, where the original scene structure is completely washed out as shown in \cref{fig:motivation}.

The fundamental reasons include 1) the data structure of 3DGS is not compatible with existing VAE models:
It is, in nature, highly unstructured due to its heterogeneous features (\textit{i.e.}, representing geometry, appearance, and lighting) and irregularity like point clouds. Unlike object-level 3DGS representations that either have high quality multi-view rendered images from synthetic data~\cite{deitke2023objaverse} or well-captured 360-degree images from real world~\cite{yu2023mvimgnet}, the hallucination in 3DGS of a general scene is prominent due to insufficient multi-view observations during per-scene optimization, as shown in \cref{fig:motivation}. Additionally, the large number of 3D Gaussians in a 3D scene makes it challenging to achieve a low-dimensional latent embedding. 2) Secondly, 3DGS representation of each scene has different global scene scales and also the individual scaling values of each 3D Gaussian primitive, making scaling up representation learning over a large number of 3DGS scene representations difficult.


We address the first problem by introducing a new 3DGS VAE module called Can3Tok, which tokenizes 3DGS inputs into canonical 3D tokens in a transformer-based VAE framework.
To embed diverse and unstructured 3DGS representations with a large number of Gaussian primitives into a compact latent space, we first employ cross-attention with a low-dimensional learnable query, encoding 3DGS representations into a small-sized tensor to enhance the efficiency of subsequent self-attention computations. The learnable query is initialized with the coordinates of a regular volume in canonical space, introducing a structured geometry prior that facilitates representation learning on irregular data.


Due to the inconsistent scale over scenes introduced by the structure-from-motion initialization, we found that even a powerful VAE model alone fails to converge on more than a thousand input scene representations. In contrast, previous works on representation learning and generation for 2D images or 3D object-level tasks do not encounter this issue, as their data either naturally has a fixed size or resolution or is consistently bounded within a uniform scale~\cite{deitke2023objaverse,chang2015shapenet}. Therefore, we propose 3DGS normalization to unify both global scene scale and three-dimensional scaling values of all 3D Gaussian primitives. To further enhance quality and mitigate the impact of 3DGS tokens from hallucinated scene regions, we filter out noisy areas in the 3DGS reconstruction caused by insufficient supervision during per-scene optimization. This is achieved through semantic segmentation guidance~\cite{kirillov2023segment} and K-nearest propagation among 3D Gaussian primitives, allowing us to extract only the cleanest and most salient scene partitions for training the VAE.
%

We validate our design using the same training and testing splits from DL3DV-10K~\cite{ling2024dl3dv}, an open-source scene-level dataset. We found that only our Can3Tok successfully converges on 3D scenes in training and generalizes well to unseen input scenes in inference, whereas other convolutional-based and transformer-based methods, with more parameters and network layers than Can3Tok, fail to converge on even a few hundred scenes in training and exhibit no generalization ability to unseen inputs in inference.
Our ablation study highlights the benefits from each proposed component for reconstruct local details.
We show that our latent space could serve as a prototype for future 3DGS generative tasks such as text-to-3DGS or image-to-3DGS generation. In summary, our main contributions include:
\begin{itemize}
    \item We propose Can3Tok, the first VAE model that tokenizes the scene-level 3DGS data into the canonical tokens by cross attention, enabling an unified latent representation learning that significantly outperforms existing VAEs;
    \item We propose a comprehensive data processing framework for 3DGS representation, including normalization to address scale inconsistency for large-scale training, as well as semantic-aware filtering and data augmentation to enhance output quality;
    \item  We showcase feedforward image-guided and text-guided scene-level 3DGS generation as applications of our Can3Tok.  
\end{itemize}

\begin{figure*}[tb]
    \centering
    \includegraphics[height=4.5cm]{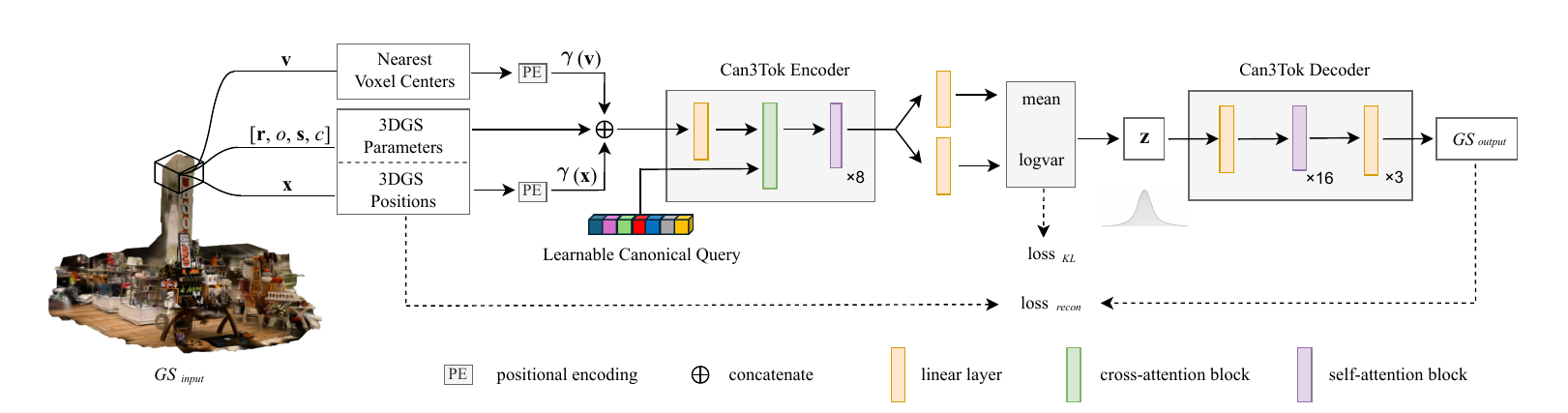}
    \vspace{-6mm}
    \caption{
        Can3Tok processes a batch of per-scene 3D Gaussians, with a batch size of B, where each scene contains the same number of Gaussians N. The encoder encodes input Gaussians into a low-dimensional latent space followed by a VAE reparametrization. And the decoder reconstructs the embeddings back into 3D space, corresponding to the original input 3D Gaussians. 
    }
    \vspace{-5mm}
    \label{fig:main}
\end{figure*}
\section{Related Work}
\paragraph{3D Gaussian Splatting.}

3D Gaussian splatting (3DGS)~\cite{kerbl20233d} has become one of the most popular 3D representations due to its flexibility and faster rendering than neural radiance fields (NeRF)~\cite{mildenhall2021nerf}. Its discrete nature can be greatly beneficial for tokenization and feed-forward 3D reconstruction with only a few or even single image inputs~\cite{zhang2025gs,szymanowicz2024splatter,szymanowicz2024flash3d}. 3DGS is akin to a point cloud and enables per-pixel parameter prediction and lifting~\cite{fan2024instantsplat,zhang2025gs} for novel-view synthesis like SynSin~\cite{wiles2020synsin}. Advancements have been specifically designed for large-scale and complex scene reconstruction~\cite{lu2024scaffold,lin2024vastgaussian,kerbl2024hierarchical,cheng2024gaussianpro}, real-time rendering~\cite{niemeyer2024radsplat} and anti-aliasing~\cite{yu2024mip,liang2024analytic}.

However, since 3DGS encompasses richer geometric and appearance features and offers greater parameter flexibility than conventional point clouds, directly integrating diffusion modules for generation is non-trivial. Recent explorations in 3D Gaussian embedding~\cite{roessle2024l3dg,ma2024shapesplat} and rearrangement~\cite{zhang2024gaussiancube} have focused on converting raw 3D Gaussian representations into more structured forms through voxelization or low-level encoding since diffusion models such as UNet and DiT cannot learn a denoising process. These approaches are currently limited to small-scale or object-level content. 
%
\vspace{-5mm}
\paragraph{Latent-Space Modeling.}

Structured latent space modeling across various data modalities, including images, videos, audio, and 3D representations, has been a long-standing problem for its advantages in compression, efficiency, and generalizability. Especially in generative tasks, the mapping between different input modalities is typically achieved through alignment in latent space, as seen in models like CLIP~\cite{radford2021learning} for text-image alignment.
Principal component analysis (PCA)~\cite{shlens2014tutorial} is widely used to compress the data in any modality into low-dimensional features by linear dimensionality reduction techniques.
To improve the efficiency and compactness, researchers have explored a non-linear approach by learning a neural network for each specific data modality:
Variational autoencoder (VAE)~\cite{kingma2013auto} is now a prototype neural architecture to project 2D visual data such as image~\cite{vahdat2020NVAE} or video~\cite{zhao2024cv} into the latent space under an encoder-decoder framework supervised by reconstruction loss and KL-divergence loss~\cite{belov2011distributions}. 
Such VAE framework is also applicable to 1D audio~\cite{luo2019learning} or 3D volume data~\cite{brock2016generative} realized by multi-layer perceptrons (MLP)~\cite{almeida2020multilayer} or 3D convolutional neural networks (3DCNN)~\cite{tran2015learning}, respectively. 
To model the latent space for the unstructured 3D point clouds, PointNet~\cite{qi2016pointnet,qi2017pointnet} and PointTransformer~\cite{zhao2021point} were two representative architectures leveraging MLP layers and attention mechanisms, respectively.
In the recent generative AI paradigm, many efforts have been made to combine all those modality in the sharable latent space, considering them as unique tokens to augment the generation quality and generation spectrum by multimodal learning~\cite{wang2022multimodal,wang2024mio}.
While 3DGS is emerging 3D representation, no prototype model for its latent space modeling has yet been actively explored where the application of existing 3D-based VAE highly suffers from the significant unstructuredness of 3DGS mentioned above. 

\vspace{-4mm}
\paragraph{Object-Level 3D Generation.}

Object-level 3D content generation has achieved significant success in many applications, \textit{e.g.}, 3D mesh generation~\cite{gao2022get3d} and texturing~\cite{michel2022text2mesh}, and point-cloud generation~\cite{nichol2022point}. With the tremendous progress made by 2D diffusion models, more diverse and high-fidelity 3D generation emerges by combining implicit 3D representations~\cite{mildenhall2021nerf} with 2D- or 3D-aware diffusion priors~\cite{rombach2022high,liu2023zero,poole2022dreamfusion,han2025vfusion3d,hu2024mvd,shi2023mvdream}. Several follow-up works~\cite{tang2023dreamgaussian,yi2023gaussiandreamer,tang2025lgm} use 3DGS as an alternative to NeRF for faster rendering. 
More impressively, transformer-based architectures~\cite{vaswani2017attention,dosovitskiy2020image,jaegle2021perceiver,jaegle2021perceiverio,hong2023lrm,jiang2023leap,jun2023shap} significantly boost the speed, scalability, and quality of 3D generation~\cite{li2023instant3d,zhao2024michelangelo,xu2023dmv3d} and token prediction~\cite{chen2024meshanything,siddiqui2024meshgpt} without any online optimization.
%
%
While impressive, all those methods are specially designed for a complete object whose application to a general 3D scene often fail due to the complexity gap between objects and scenes e.g. scale-inconsistency and partial observations, which lead researchers to explore the generative framework uniquely designed for a general scene.  

%

\vspace{-5mm}
\paragraph{Scene-Level 3D Generation.}

Scene-level 3D generation usually refers to large-scale or even unbounded 3D content creation. One exceptional example is perpetual view generation~\cite{liu2021infinite,li2022infinitenature}, which allows for view-consistent video or 3D generation with an arbitrary long camera trajectory. Besides these auto-regressive approaches, some efforts~\cite{chung2023luciddreamer,lin2023infinicity,fridman2024scenescape,yu2024wonderworld,yu2024wonderjourney,hollein2023text2room}, enhance global consistency by directly incorporating various 2.5D or 3D cues such as depth, normal, and point cloud into 3D representations such as meshes, NeRFs, or 3DGS for rendering. Though impressive, these methods mostly inherit cumbersome optimization in an iterative manner, 
Moreover, all recent advancements in scene-level 3D generation rely on 2D supervision, either from images or 2D diffusion models, making feed-forward generation unfeasible without a low-dimensional embedding for large-scale 3D representations. In this work, we propose Can3Tok for scalable 3DGS embedding that be directly combined with widely-used diffusion architectures for feed-forward 3D generation without image-space supervision.
\section{Our Approach}
\subsection{3DGS Preliminaries}

Given multi-view images with associated camera parameters, a 3D scene depicted in the images can be represented by a set of elliptical 3D primitives, each with an internal radiance field that follows a Gaussian distribution. An individual 3D Gaussian primitive has several parameters, including its 3D center $\mathbf{x}\in\mathbb{R}^3$, rotation represented by a quaternion $\mathbf{r}\in\mathbb{R}^4$, opacity $o\in\mathbb{R}^1$, scaling $\mathbf{s}\in\mathbb{R}^3$, view-independent RGB color $c\in\mathbb{R}^3$, and view-dependent color with high-dimensional spherical harmonics $c_h \in \mathbb{R}^h$.

\subsection{Can3Tok Design}

Can3Tok is a transformer-based VAE architecture comprising an encoder and a decoder.
The encoder maps the tokenized 3DGS to a latent space, and the the decoder reconstructs the original input 3DGS data. 
The design of our VAE model is illustrated in \cref{fig:main}. 


\vspace{-5mm}
\paragraph{Encoder.}

The inputs to the encoder is a set of 3DGS data.
Similarly to the positional encoding in NeRF~\cite{mildenhall2020nerf}, we apply Fourier positional encoding on 3D Gaussian centers $\gamma(\mathbf{x}):\mathbb{R}^{N\times3} \rightarrow \mathbb{R}^{N\times L_B}$ with a pre-set maximum
band $L_B$ to better capture high-frequency components in a low-dimensional embedding. 
While theory and experiments~\cite{tancik2020fourier, mildenhall2021nerf} have validated the effectiveness of this positional encoding approach on MLP-based networks, it has also proven effective across a broader range of architectures including transformers~\cite{jaegle2021perceiverio,siddiqui2024meshgpt,chen2024meshanything,zhao2024michelangelo,jaegle2021perceiver}.
Since the 3DGS representation is unstructured, we append more structured ``anchors'' as representative locations to each 3D Gaussian, to reduce the burden of the encoder representation learning:
We build a volume on the space of 3D Gaussians with resolution $V^3$ and apply the same Fourier positional encoding on the voxel center nearest to each 3DGS position $\gamma(\mathbf{v}):\mathbb{R}^{N\times 3} \rightarrow \mathbb{R}^{N\times L_B}$. Please also refers to our supplementary materials for the effect of nearest voxel coordinates appending. While appending voxel coordinates is not crucial to Can3Tok's success, we find that it enhances output quality, particularly in level-of-detail.

The encoder takes per-scene 3DGS information $\mathcal{G}\in\mathbb{R}^{N\times (2\times L_B+C)}$ including $\gamma(\mathbf{x})$, $\gamma(\mathbf{v})$, and other 3D Gaussian parameters, where $N$ is the number of Gaussians per each scene and $C$ is the size of 3DGS feature dimension.
The encoder starts with a linear layer that maps the inputs into $key$ and $value$, and they are tokenized by a cross-attention that takes in the $key$, $value$, and a learnable canonical $query$, inspired by PerceiverIO~\cite{jaegle2021perceiverio}. This is because an input scene has more than 10k 3D Gaussians, making the naive self-attention computationally expensive.
Importantly, the canonical query is initialized with regular voxel grids and associated descriptors, \textit{i.e.,} $query\in\mathbb{R}^{M\times (P+Q)}$ where $M$ is the number of canonical voxels, $P$ denotes voxel's position, and $Q$ is the size of the descriptors, and further optimized during the training of Can3Tok.
The subsequent 8 blocks of self-attention are applied to the tokens to explore the affinities or any other relationships among latents while preserving its dimension unchanged.


\paragraph{Latent space.}

Following the common VAE design~\cite{kingma2013auto}, the outputs from our encoder are projected into two latent vectors representing mean $\mu$ and log-variance $\log\boldsymbol{\sigma}^2$. The corresponding embedding $\mathbf{z}$ is sampled with the VAE reparameterization trick:
\begin{equation}
    \label{eq:reparamerization}
    \mathbf{z} = \boldsymbol{\mu} + \boldsymbol{\epsilon}*\exp(0.5*\log\boldsymbol{\sigma}^2)
\end{equation}
where $\boldsymbol{\epsilon}~\sim\mathcal{N}(\bf{0},\bf{I})$ is sampled from a normal distribution. 

\paragraph{Decoder.}

As shown in \cref{fig:main}, the decoder takes latent samples $\mathbf{z}$ and recovers 3DGS parameters $GS_\text{output}$ through a linear layers and 16 blocks of self-attention. Unlike PerceiverIO~\cite{jaegle2021perceiver}, which is designed for the prediction on a discrete target domain, our decoder does not have a pre-defined output query as we aim for the reconstruction in continuous 3D space. To this end, the decoder's tail includes multiple linear layers with non-linear activation function for mapping a latent space into 3D continuum. Although the multi-layer perceptron at the end of the Can3Tok decoder has only a limited number of learnable parameters, both its inputs and outputs are within a bounded space, making the latent-to-3D mapping feasible and less computationally burdensome.

\paragraph{Training Objective.}

We follow the common setting of training a VAE~\cite{kingma2013auto}. The objective for our model optimization is to minimize the following loss:
\begin{equation}
 \mathcal{L} = \mathrm{Dist}(GS_\text{output}, GS_\text{input}) + \lambda\mathcal{L}_\mathrm{KL}(\mathbf{z}, \mathcal{N}(\bf{0},\bf{I}))
  \label{eq:loss}
\end{equation}
where $GS_\text{input}$ are input 3D Gaussian representations and the scalar $\lambda$ balances two losses:
$\mathrm{Dist}$ measures the $L2$ distance between the recovered 3DGS and the ground truth 3DGS across all different feature channels, and $\mathcal{L}_\mathrm{KL}$ is the KL divergence between the latent space $\mathbf{z}$ and a normal distribution $\mathcal{N}(\bf{0},\bf{I})$ so as to have a structured distribution.


\subsection{3DGS Processing}
\label{data_process}
A fundamental challenge in scaling up VAE models lies in the scale inconsistency across different 3DGS scene representations as we found that none of the existing methods, including our Can3Tok, can generalize well to a large number of scene-level 3DGS representations. Since neither global scene scales or scaling factors of each 3D Gaussian primitive are not metric as the using of COLMAP~\cite{schoenberger2016sfm} for camera pose estimation and 3D SfM point triangulation, we propose to unify 3DGS representations into a bounded scale for scaling up training and semantic-aware filtering for improving the reconstruction quality of model outputs. 

\begin{figure}[t]
    \centering
    \begin{minipage}[t]{1.0\linewidth} 
        \centering
        \includegraphics[width=\linewidth]{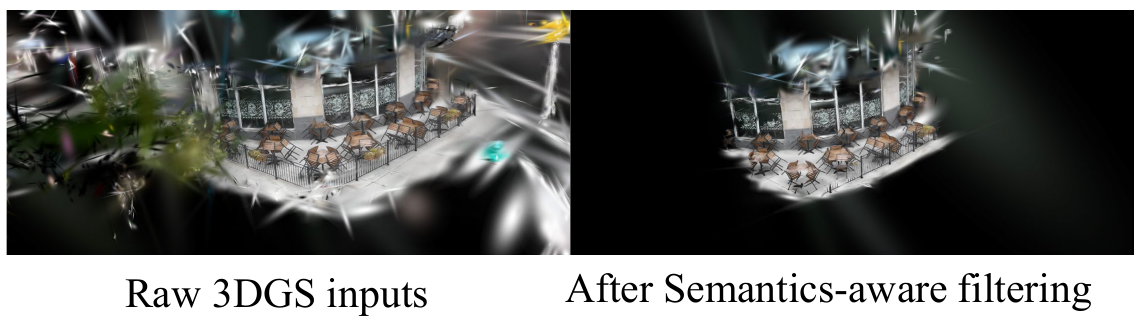} 
    \end{minipage}
    \vspace{-6mm}
    \caption{
        Before and after our semantic-aware 3DGS filtering.
    }
    \label{fig:filtering_comp}
    \vspace{-6mm}
\end{figure}
\vspace{-3mm}
\paragraph{Normalization.} 
Since there are no established techniques for scalable 3D Gaussian representation learning, we take inspiration from 2D image representation learning~\cite{malo2003normalized,wang2020attentive}, where the size of all input images are the same and their RGB channels are normalized into a bounded coherent scale (\textit{e.g.}, [-1,1]) has proven effective for accelerating model convergence and generalization. Specifically, to apply normalization to 3DGS data,
we mean-shift the 3DGS centers $\textbf{x}$ to the origin of world space, and bound all 3DGS into a sphere with radius $r$ while we also re-scale the scaling factor $\textbf{s}$ of each 3DGS to become $\hat{\mathbf{s}}$:
\begin{align}
    \mathrm{translate} &= - \frac{1}{n}\sum^{n}_{i=1}\mathbf{x}_i, \nonumber\\
    \mathrm{scale} &= \frac{r}{\max|\mathbf{x} + \mathrm{translate}|_2*1.1}, \nonumber\\
    \hat{\mathbf{x}} &= (\mathbf{x} + \mathrm{translate}) * \mathrm{scale}, \nonumber\\
    \hat{\mathbf{s}} &= \mathbf{s} * \mathrm{scale}, \nonumber\\
    \hat{\textbf{T}}_i &= (\textbf{T}_i + \mathrm{translate}) * \mathrm{scale},
    \label{eq:normalization}
\end{align}
where $\textbf{T}_i$ and $\hat{\textbf{T}}_i$ are center locations of cameras that associate to the training views of 3DGS scene representations. Unlike the RGB channels of 2D images, we preserve all other 3DGS attributes during normalization, as they are heterogeneous and retain their physical meaning only within their original numerical range. The intuition behind Eq.~\ref{eq:normalization} is that: If the same transformation is applied to the camera centers associated with the same 3DGS representations, transforming $\textbf{T}_i$ into $\hat{\textbf{T}}_i$ while keeping the camera orientation unchanged, then images rendered from the 3DGS before normalization, using the original camera view with center position $\textbf{T}_i$, will be identical to those rendered from the 3DGS after normalization, using the same camera view but with the re-scaled position $\hat{\textbf{T}}_i$. Therefore, another benefit of the proposed 3DGS normalization is that we can recover the metric scales of scenes by estimating metric depth on images rendered from our generated outputs using depth foundation models.


\vspace{-3mm}
\paragraph{Semantic-aware Filtering.}

\begin{figure}[t]
    \centering
    \begin{minipage}[t]{1.0\linewidth} 
        \centering
        \includegraphics[width=\linewidth]{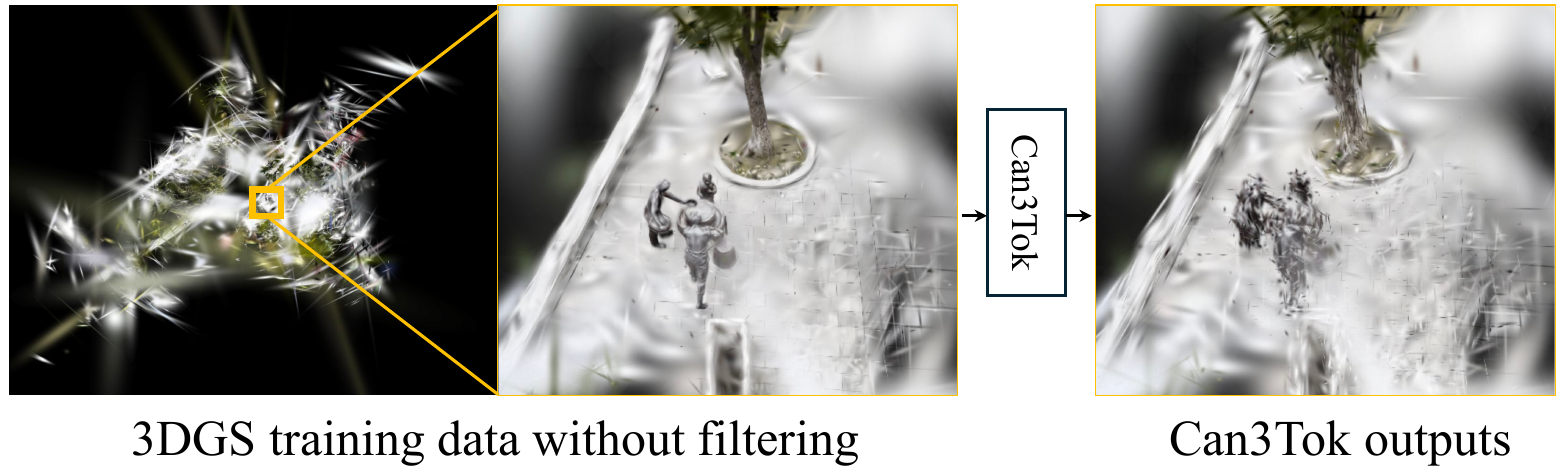} 
    \end{minipage}
    \vspace{-7mm}
    \caption{
        Noise-dominated 3DGS training data hurts the latent-space modeling of Can3Tok where meaningful local details largely collapse, which motivate us to have clean scene-level 3DGS data with semantics-aware filtering.
    }
    \vspace{-6mm}
    \label{fig:global_local}
\end{figure}

3DGS reconstructions from general scenes often contain noise artifacts like floaters due to the lack of visual observations (unlike objects which are normally captured with sufficient views).
Though Can3Tok can effectively compress 3DGS inputs as latent representations, we experimentally found that such noise deteriorates the latent representation where high-frequency details are washed out in the decoded 3DGS output as shown in \cref{fig:global_local}.
To address this issue, we apply semantic-guided filtering to the raw 3DGS input to subsample as-clean-as-possible 3DGS primitives.
Specifically, we apply LangSam~\cite{kirillov2023segment}, a text-guided variant of segment anything model, on the middle frame of each scene video with the text prompt ``the most salient region''. LangSam crops out the most semantically meaningful region depicted in the image. We pick one Gaussian within the segmentation mask and incrementally include more Gaussians based on a K-NN algorithm in 3D space until reaching a pre-set number $N$. As shown in \cref{fig:filtering_comp}, such semantic filtering can preserve the most semantically meaningful contents while removing the less salient and noisy Gaussians.

\section{Experiments}

\subsection{Implementation Details}
\textbf{Dataset}: We run 3DGS on all videos from DL3DV-10K dataset~\cite{ling2024dl3dv} (with 6:1 training/testing split), where camera positions and SfM points are obtained by COLMAP~\cite{schonberger2016structure} for 3DGS initialization. 
%
%
We set $N=100K$ for each scene representation by applying an upper bound on Gaussian densification and pruning during per-scene optimization.\\
\textbf{Data Augmentation}: We apply random SO(3) rotations to input 3DGS representations to get more $GS_{input}$ as a way of data augmentation, similarly to the common random-rotation augmentation on 2D images. \\
\textbf{Architecture}: We implement our encoder with 1 linear layer, 1 cross-attention block, 8 self-attention blocks, and 2 linear layers for mapping latents at the bottleneck into mean and log-variance. Our decoder starts with one linear layer followed by 16 self-attention blocks and ends with 3 linear layers. Self- and cross-attention blocks are of multi-head with 12 heads and 64 dimensions each, implemented using Flash-Attention~\cite{dao2022flashattention}. Layer normalization is appended to each linear layer and attention block. Latent query has size $\mathbf{Q}\in\mathbb{R}^{256\times 768}$. Mean $\mu$, log-variance $\log\sigma^2$, and $\mathbf{z}$ are in $\in\mathbb{R}^{64\times 64\times4}$, which has exactly the same size as the latent space of Stable Diffusion~\cite{rombach2021highresolution}. The output from Fourier positional encoding has size $L_B=51$, specifically $\gamma(\mathbf{x}): \mathbb{R}^{N\times 3}\rightarrow \mathbb{R}^{N\times 51}$. Input volume has resolution $V=40$. We set the loss hyper-parameters to $\lambda=1\times10^{-6}$. Each scene has the same number $N=40K$ of Gaussians after semantics-aware filtering. We train our model on 8 A100 GPUs for 5 days. A single forward pass for encoding and decoding an input 3D scene with our model takes only $\sim$0.06\,sec, making it compatible with a diffusion module for feedforward generation, as we demonstrate later.

\begin{figure*}[t]
    \centering
    \begin{minipage}[t]{1.0\linewidth} 
        \centering
        \includegraphics[width=\linewidth]{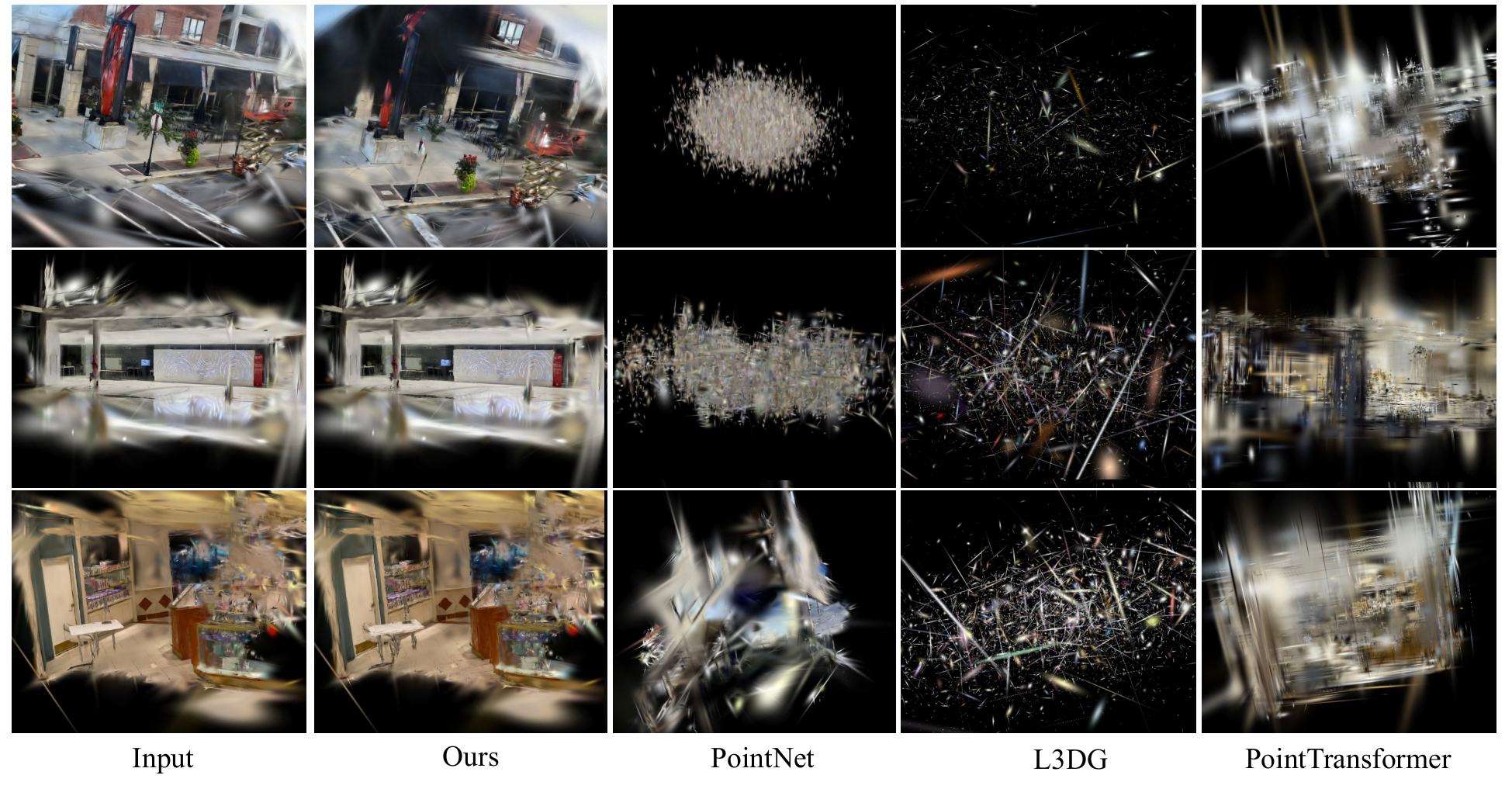} \\ 
    \end{minipage}
    \hspace{0.1\textwidth} 
    \vspace{-7mm}
    \caption{
        Qualitative comparisons between ours and other VAE outputs. Results are not cherry picked as all compared methods show zero generalization ability on novel scene inputs. PointNet and L3DG do not converge on training set.
    }
    \vspace{-6mm}
    \label{fig:unseen2}
\end{figure*}
\begin{table}[t]
    \centering
    \vspace{-3mm}
    \caption{
        Quantitative comparison on DL3DV-10K testing set (with filtering). $L_2$ error measures the distance between each pair of $GS_\text{output}$ and $GS_\text{input}$ over the test set. Failure rate is the percentage of cases where the model completely fails to reconstruct the input 3D Gaussians.
    }
    \vspace{-2mm}
        \begin{tabular}{l|cc}
        \toprule
        \multicolumn{1}{l|}{} & $L_2$ error $\downarrow$ & Failure rate$\downarrow$   \\
        \toprule
        L3DG~\cite{roessle2024l3dg} & 1200.4 & 100\% \\
        PointNet VAE~\cite{qi2017pointnet} & 1823.0 & 100\%\\
        PointTtransformer~\cite{zhao2021point}&  230.7 & 70\%\\
        Ours &  \bf{30.1} & \bf{2.5}\% \\
        \bottomrule
        \end{tabular}
    \label{tab:metric}
    \vspace{-2mm}
\end{table}
\begin{table}[t]
    \centering
    \caption{
        Quantitative comparisons and ablation studies on DL3DV-10K testing set. $L_2$ error measures the distance between each pair of $GS_\text{output}$ and $GS_\text{input}$ over the test set. Failure rate is the percentage of cases where the model completely fails to reconstruct the input 3D Gaussians.
    }
        \begin{tabular}{l|cc}
        \toprule
        \multicolumn{1}{l|}{} & $L_2$ error $\downarrow$ & Failure rate$\downarrow$   \\
        \toprule
        Ours (w/o Learnable Query) & $10^{25}$ & 100\% \\
        Ours (w/o normalization) & 1889.7 & 100\%\\
        Ours (w/o voxel appending) & 50.5 & 4.3\% \\        
        Ours (w/o data filtering) & 73.3  & 6.1\%\\    
        Ours (w/o data augmentation) & 53.3 & 4.6\% \\
        \midrule
        Ours (full) &  \bf{30.1} & \bf{2.5}\% \\
        \bottomrule
        \end{tabular}
        \vspace{-5mm}
    \label{tab:quanti}
\end{table}

\begin{figure}[ht]
    \centering
    \begin{minipage}[t]{1.0\linewidth} 
        \centering
        \includegraphics[width=\linewidth]{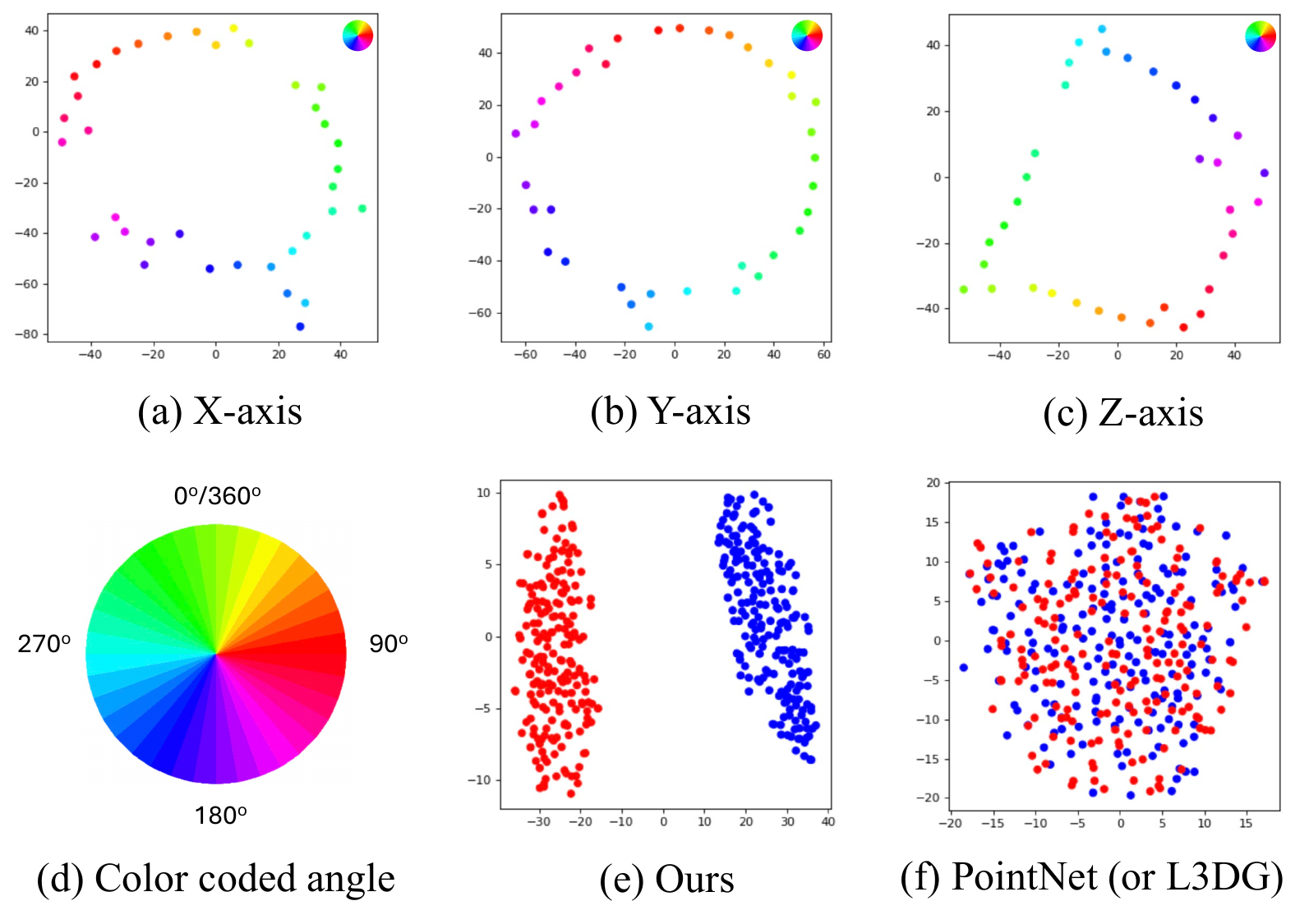} 
    \end{minipage}
    \vspace{-6mm}
    \caption{
        t-SNE visualizations of the latent space of 3DGS for the same scene with 36 linearly interpolated SO(3) rotations from 0 to 360 degrees. All three rotations exhibit patterns of closed loops, demonstrating that our model preserves spatial information in the latent representations.  In (e) and (f), \textcolor{red}{red} dots are latent embeddings of the same scene but with 200 random SO(3) rotations and  \textcolor{blue}{blue} dots are latent embeddings of different scenes.
    }
        \vspace{-11mm}
    \label{fig:tsne_2}
\end{figure}
\subsection{Baselines and Metrics}

We compare our VAE model to existing 3D-based VAE models.
L3DG~\cite{roessle2024l3dg} is a recent method for object-level 3DGS encoding and decoding. We implement L3DG and other convolution-based architectures with Minkowski Engine~\cite{choy20194d} and spconv~\cite{spconv2022} following L3DG paper. We also compare with a PointNet-based autoencoder~\cite{qi2017pointnet} with more network layers to increase its model capacity. Another baseline method is PointTransformer~\cite{zhao2021point}, which is a popular transformer-based architecture. All compared methods are trained and tested on the same train and test sets from DL3DV-10K~\cite{ling2024dl3dv} with the same data processing as training our Can3Tok, to evaluate if they can accurately generalize to unseen input 3DGS. Since 3D Gaussians have more parameters than simple location coordinates, we use average $L_2$-norm across all 3DGS features between each pair of $GS_\text{output}$ and $GS_\text{input}$ on the test set as an evaluation metric. Moreover, we measure the failure rate, defined as the percentage of cases in which the model completely fails to reconstruct the input 3D Gaussians, as shown in \cref{fig:unseen2} and more qualitative results in the supplementary material.

\subsection{Results}

In our experiments, PointNet and L3DG cannot even converge with more than 500 3D scenes while PointTransformer performs slightly better but still produces poor visual quality. As shown in \cref{tab:metric}, these methods fail to generalize to unseen 3D scenes with 100\% failure rate, while our design shows great potential for its scalability with low $L_2$ error and failure rate. The failure rate is defined as the percentage of scenes whose $L_2$ error of reconstruction exceeds 1000.0, which are completely not recognizable as shown in \cref{fig:unseen2}. Our model successfully reconstructs original input 3DGS, and neither convolution- or transformer-based methods are able to succeed in decoding the 3DGS scenes by losing the original global shape and local details. While PointTransformer~\cite{zhao2021point} outperforms L3DG and PointNet in terms of $L_2$ error and failure rate, the reconstructions show stretched and distorted patterns on all test samples. Qualitative results shown in \cref{fig:unseen2} and supplementary materials are general and not cherry picked. This failure mode observed in convolution-based VAEs is also highlighted in the concurrent work Bolt3D~\cite{szymanowicz2025bolt3d}.


\subsection{Latent-Space Analysis}

\begin{figure}[ht]
    \centering
    \begin{minipage}[t]{1.0\linewidth} 
        \centering
        \includegraphics[width=\linewidth]{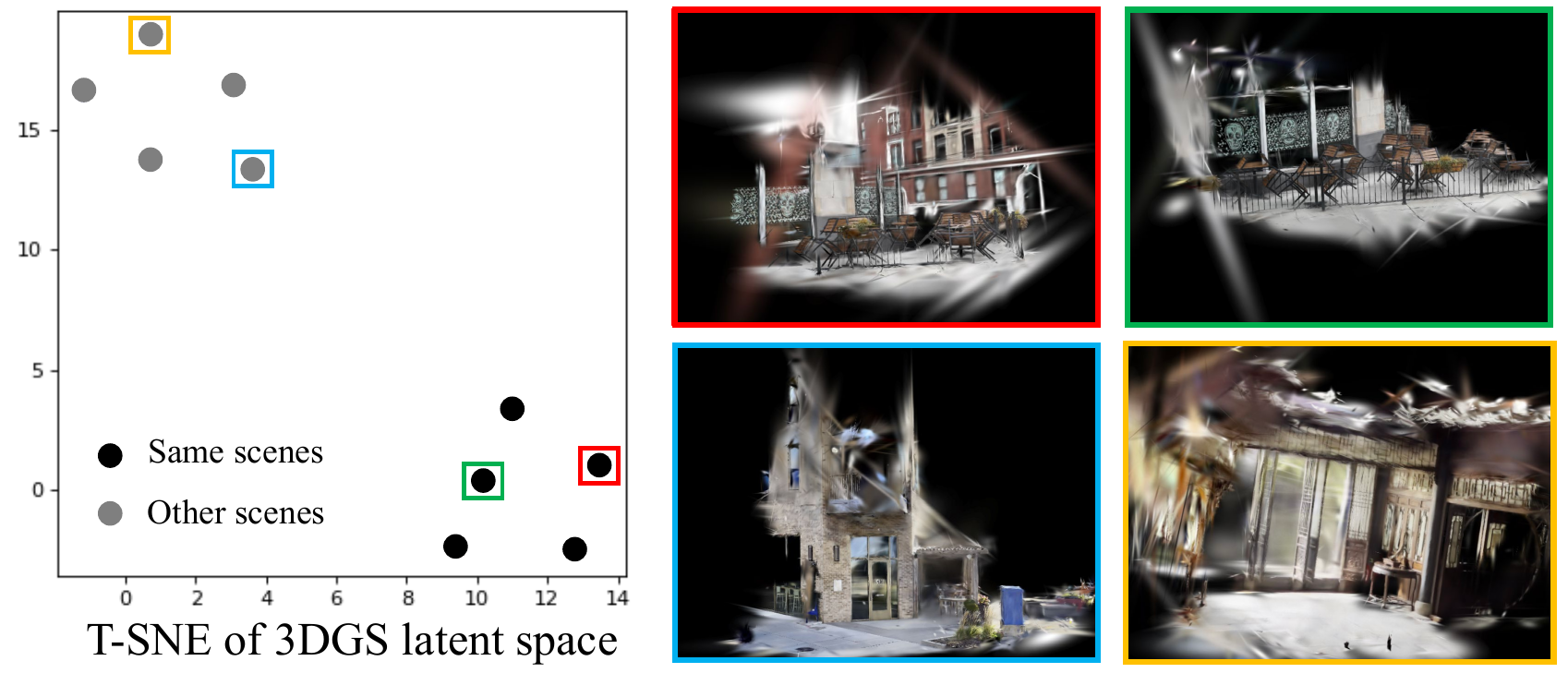} \\ 
    \end{minipage}
    \vspace{-5mm}
    \caption{
         Given 3DGS reconstruction for various scenes, we randomly subsample 3DGS similar to the semantics-aware filtering in \cref{data_process}; and demonstrate the t-SNE of their latent space. The latent spaces from same scenes, visualized in \textcolor{red}{red} and  \textcolor{green}{green} images, are closer each other, and otherwise for other scenes.}
         \vspace{-5mm}
    \label{fig:tsne_dist}
\end{figure}

It remains inconclusive why our proposed method genuinely captures structured and meaningful 3D geometric patterns within the representation instead of merely memorizes the input 3DGS. This question cannot be fully answered by quantitative metrics and qualitative comparisons in spatial domain.\\
\textbf{Spatial Encoding}: Therefore, we highlight our structured latent space by exploring the spatial relationship between inputs and the associated latent embeddings via t-SNE~\cite{van2008visualizing} visualization shown in \cref{fig:tsne_2}. We encode a 3DGS scene with different SO(3) rotations into latent embeddings in inference stage. Even though no explicit constraints on latent space were applied during training, our method automatically discovers the spatial correlation between inputs and latent embeddings. While other baseline methods mix up the same scene under different 3D orientations and other scenes in latent space. Therefore the decoder is fail to correctly decode latents back to different 3DGS representations.\\
\textbf{Semantic Encoding}: We also highlight the ability of our latent representation to abstract semantic information of inputs instead of merely memorizing all 3D Gaussians. In \cref{fig:tsne_dist}, subfigures are two different 3D Gaussian filterings or croppings from raw 3D Gaussians but covering the same 3D contents (some outdoor chairs and desks on the ground together with a wall), while two samples are from a completely different scene. In latent space, latent embeddings from the same scenes are close to each other (black dots), and along with other different scenes (gray dots) are far away. Additionally, the different SO(3) rotations of the same scene can be close to each other in latent space as shown in \cref{fig:tsne_2} as they are semantically similar. 


\begin{figure*}[t]
\centering
    \begin{minipage}[t]{1.0\linewidth} 
        \centering
        \includegraphics[width=\linewidth]{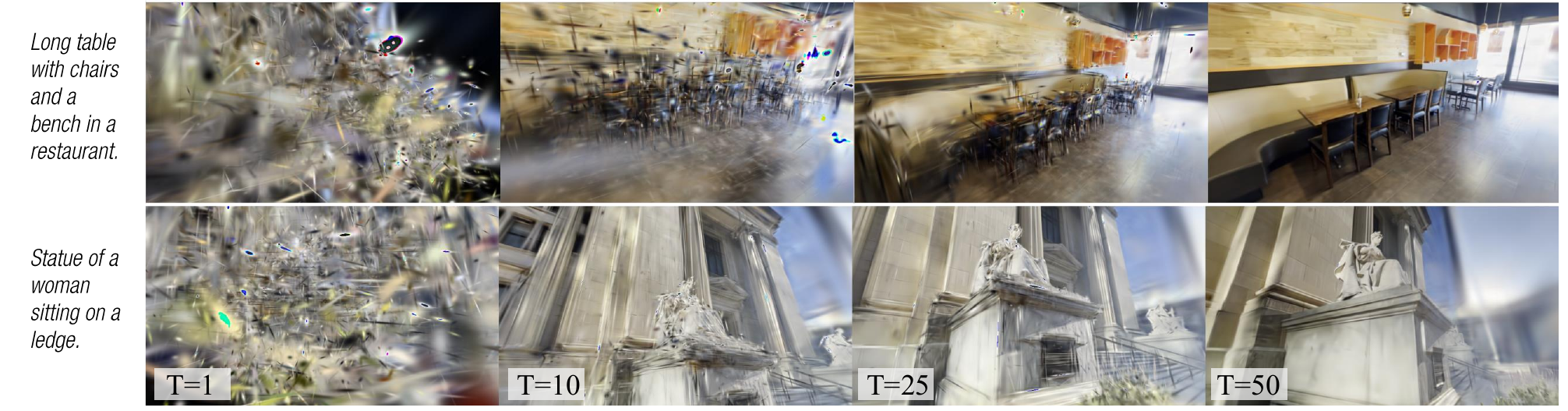} \\ 
    \end{minipage}
        \vspace{-5mm}
    \caption{
        Two examples of text-conditioned generation with a latent diffusion UNet, which is trained on our 3DGS embeddings of scenes and corresponding text prompts. The scene labels are shown in the captions, and the images are rendered with camera positions around the generated 3DGS by denoising the latents $\mathbf{z}$ passing through our Can3Tok decoder. $\rm T$ denotes the denoising time step. Please note that, while the images are rendered from a fixed camera viewpoint, the global 3DG structures are shifting during the denoising process, which lead to the viewpoint-shifting sensation. 
    }
    \vspace{-4mm}
    \label{res_diffusion}
\end{figure*}

\begin{figure}[t]
\centering
    \begin{minipage}[t]{1.0\linewidth} 
        \centering
        \includegraphics[width=\linewidth]{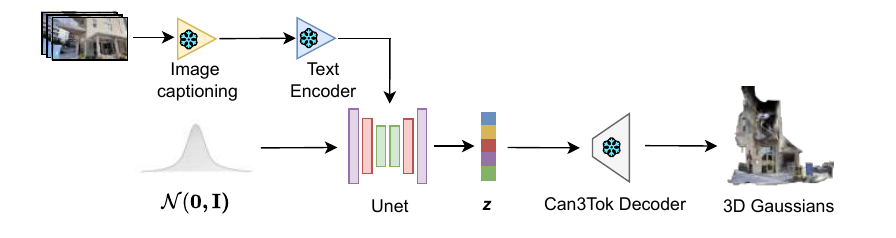} \\ 
    \end{minipage}
    \vspace{-5mm}
    \caption{
        A pipeline for text-conditioned latent diffusion model for 3D Gaussian generation using Can3Tok latent embeddings. The pretrained image-captioning, text-encoder and Can3Tok decoder modules are frozen during training the denoising UNet model. Snow labels denote the pre-trained models.
    }
    \vspace{-5mm}
    \label{fig:text-conditioned}
\end{figure}
\subsection{Ablation Study}
We study the effectiveness of each proposed module and technique. Please refer to \cref{tab:quanti} for the quantitative ablations. Also, we describe the qualitative results and the detailed settings of ablation study in the supplementary materials, which include 1) without learnable query, 2) no 3DGS normalization, 3) no 3DGS data filtering, 4) no voxel coordinates appending, 5) no data augmentation. 

In \cref{fig:global_local}, We demonstrate that semantic-aware data filtering is essential for capturing high-frequency details of 3DGS in the latent space, as it mitigates the negative impact of training a VAE model with imperfect 3DGS reconstruction results. Since learning raw 3DGS with floaters and noise deteriorates the high-frequency correlation between the input 3DGS and the latent space. Consequently, without the proposed data filtering, $L_2$ increases, as shown in ~\cref{tab:quanti}. More importantly, while Can3Tok is the only method capable of converging on thousands of training samples, it fails to generalize to unseen test data without 3DGS normalization (w/o normalization). This reveals that a well-designed model alone cannot address the unique scale-inconsistency issue in scene-level 3D representations.

\vspace{-1mm}
\subsection{Application}

We showcase that our 3DGS latent embedding could be useful for various generative tasks including text-guided and image-guided 3DGS generations.

\vspace{-4mm}
\paragraph{Text-to-3DGS Generation.}

As shown in \cref{fig:text-conditioned}, we train a diffusion UNet for mapping a noise vector sampled from a normal distribution into a meaningful 3D Gaussian embedding $\mathbf{z}$, conditioned on text prompts. Since the DL3DV-10K dataset lacks labels or scene text descriptions for scenes, we caption the middle frame of each video to label each scene where we use a pre-trained BLIP model~\cite{li2022blip} whose output text prompts are consistently concise and capture salient semantic information. After this labeling, we train the diffusion model on $(\text{text label}, \mathbf{z})$ pairs. At inference, the UNet samples a $\mathbf{z}_T \sim \mathcal{N}(\bf{0},I)$ from normal distribution and tries to denoise it with $T$ steps to approach $\mathbf{z}_0 \equiv \mathbf{z}$, conditioned on a text scene label. We showcase the denoising process of two inference examples in \cref{res_diffusion}. 


\vspace{-5mm}
\paragraph{Image-to-3DGS Generation.}

In the supplementary document, we also showcase that our 3DGS latent embedding can be combined with existing image regression modules~\cite{chadebec2022pythae} to enable image-guided 3DGS generation. To this end, we train the encoder that regresses the image to our 3DGS latent space, and we use Can3Tok decoder to construct the associated 3DGS outputs. Please refer to the supplementary for the visual results.
\subsection{Limitations}
As shown in \ref{tab:quanti}, our method did not achieve a 100\% success rate. This is due to some low-quality 3DGS reconstructions in the training set. We observed that some videos used for 3DGS reconstruction suffer from severe motion blur and an imbalanced distribution of close-up and distant views during data capture. As a result, the corresponding latent representations become less discriminative and tend to mix with those of other 3D scenes. Besides, our method is limited to the 3DGS representation, as it is more discrete and suited for tokenization than other neural representations.

%
%
\vspace{-1mm}
\section{Conclusion}

We introduce Can3Tok, the first method for scene-level 3DGS latent representations, demonstrating that all existing approaches fail without a model design and proper 3D data normalization specifically tailored to the 3DGS representation. Through latent space analysis, qualitative and quantitative comparisons, we show that our method significantly outperforms existing 3D VAE models. Additionally, we propose a 3DGS data processing approach to address the open problem of scale inconsistency in 3D representations. To further enhance quality and scalability, we introduce semantic-aware filtering and data augmentation. Finally, we showcase the practical utility of Can3Tok in 3D generative applications, including text-guided and image-guided 3D generation with 3DGS representation.

\section{Acknowledgment}
We thank Kai Zhang and Zexiang Xu for the insightful discussion.

{
    \small
    \bibliographystyle{ieeenat_fullname}
    \bibliography{main}
}

\newpage
    \twocolumn[{%
        \renewcommand\twocolumn[1][]{#1}%
        \begin{center}
            \centering
            \LARGE \textbf{\appendixname}
            \vspace{30pt}
        \end{center}%
    }]



\maketitle

\maketitle
In this supplementary material, we demonstrate the qualitative ablation study of our Can3Tok-based VAE model; more qualitative results; image-to-3DGS application and more discussions. \\


\begin{figure}[ht]
\vspace{-3mm}
    \centering
    \begin{minipage}[t]{1.0\linewidth} 
        \centering
        \includegraphics[width=\linewidth]{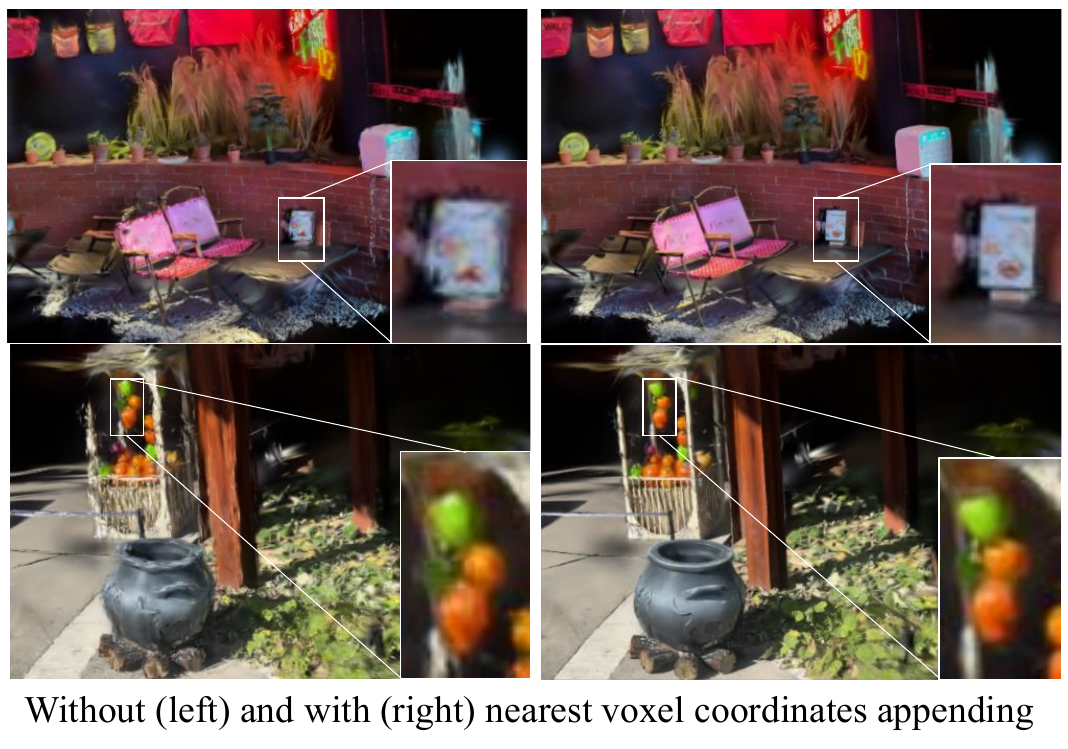} \\ 
    \end{minipage}
    \vspace{-6mm}
    \caption{Ablation study for with and without nearest voxel coordinate appending with each input 3DGS. The results indicate that appending structured volume coordinates to the unstructured input 3D Gaussians leads to better reconstruction.
    }
    \label{fig:voxel_ablation}
    \vspace{-7mm}
\end{figure}

\section{Ablation Studies}
We perform ablation study for the importance of each module of our method. In Tab. 2 of the main paper, we describe the overall quantitative comparison over different ablation studies. More specifically, we verify the performance by removing each of the following modules:

\noindent 1) \textit{w/o Learnable Query}: We remove the learnable canonical latent query and we replace the cross-attention block with self-attention. We observe that simply replacing it with self-attention fails to converge and is more likely to encounter out-of-memory issue, even with a batch size of one, although we are using a GPU with a large memory capacity (\textit{e.g.,} 80GB). In fact, this highlights the importance of our Can3Tok module for its computational efficiency. \cref{fig:can3tok_ablation} highlights that it was not possible to make a structured latent space to recover the original inputs without cross-attention with a low-dimensional learnable query.

\noindent 2) \textit{w/o normalization}: we do not apply the normalization of data to the entire 3DGS training dataset. Both Tab.2 and \cref{fig:norm_ablation} highlight the severe scale inconsistency issue if a VAE model is trained on raw 3DGS input, which hinders scaling up training across thousands of scenes. A uniform data normalization strategy is essential to allow large-scale training and improve generalization.


\noindent 3) \textit{ w/o data filtering}: we use raw 3DGS reconstruction results as a training set without data filtering. \cref{fig:filtering_ablation} implies that by suppressing the significant noise by data filtering, the models better learn the mapping between the latent and inputs in a way that preserves the local details.

\noindent 4) \textit{w/o voxel coordinate appending}: we turn off the dual positional embedding from 3DGS's position and its nearest voxel center. Instead, we append the positional embedding only from 3DGS's position. \cref{fig:voxel_ablation} show the effect of voxel coordinate appending where its to preserve the high-frequency local details in the reconstructed 3DGS.

\noindent 5) \textit{ w/o voxel data enhancement}: we disable data enhancement during training.


\section{More Results}
\vspace{-1mm}
In \cref{fig:more_quali}, we demonstrate more results from our Can3Tok with various test scenes.
\vspace{-2mm}
\section{Application: Image-to-3DGS}
\begin{figure}[b]
    \centering
    \begin{minipage}[t]{0.7\linewidth} 
        \centering
        \includegraphics[width=\linewidth]{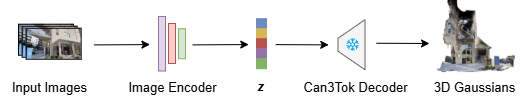} \\ 
    \end{minipage}
    \caption{Illustration of image-to-3DGS architecture.
    }
    \label{fig:image2latent_qual}
\end{figure}

\begin{figure}[ht]
    \centering
    \begin{minipage}[t]{1\linewidth} 
        \centering
        \includegraphics[width=\linewidth]{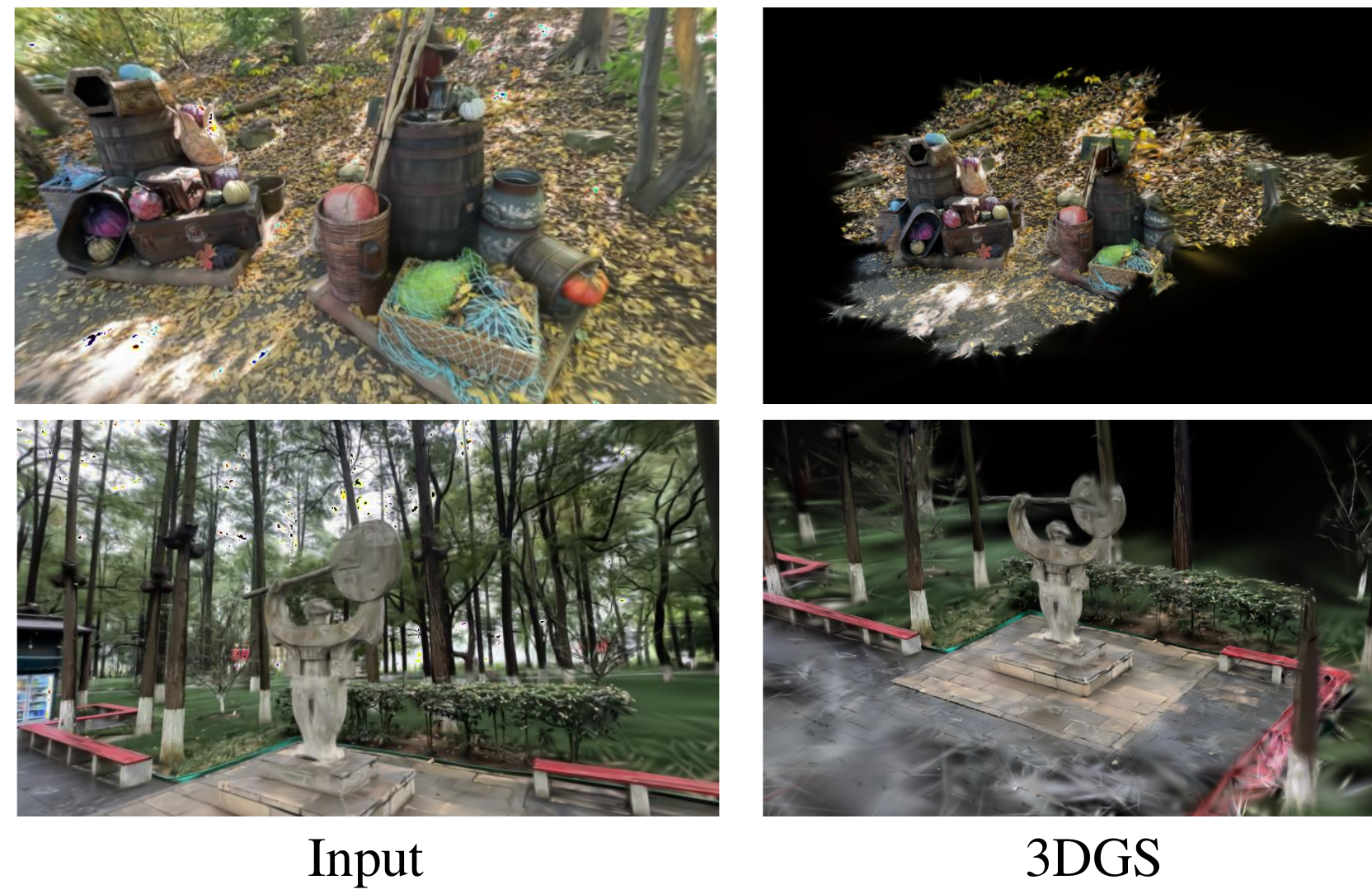} \\ 
    \end{minipage}
    \caption{Qualitative examples of our image-to-3DGS applications.
    }
    \label{fig:image2latent}
\end{figure}

In this section, we showcase the application of our Can3Tok latent space modeling.
Other than text-to-3DGS, our latent features can be used for image-to-3DGS generation. To this end, we use an image encoder (\textit{e.g.,} \cite{chadebec2022pythae}) that takes as input 2D images and outputs corresponding latents; and our pretrained Can3Tok decoder constructs the associated 3DGS scene. The pipeline is shown in \cref{fig:image2latent}. The objective of the encoder training is to minimize the $L_2$ error between the predicted latents $\mathbf{z}$ and ``Ground-Truth 3D Gaussian latents" $\mathbf{z}_{GT}$, which are obtained by inputting 3D Gaussians into the Can3Tok encoder. The reason we use a regression objective instead of a diffusion objective is that almost all text-conditioned generations including text-to-image ~\cite{rombach2022high} and text-to-(3D object)~\cite{zhao2024michelangelo} use diffusion objective due to their probabilistic nature. While image-to-3D is inherently more deterministic than text-to-3D, given that methods like Flash3D~\cite{szymanowicz2024flash3d} and SplatterImage~\cite{szymanowicz2024splatter} both use regression objectives. Therefore, we follow the similar idea: regress \textbf{z} from an input image and then predict 3DGS from \textbf{z}. We showcase some qualitative examples of this Image-to-3DGS applications in \cref{fig:image2latent_qual}. 


\section{More Discussions}
About speed, our Can3Tok VAE (1.1 s/iters) is 10 times faster than L3DG (11.3 s/iters). This is because our method accelerate self-attention steps by reducing the input dimension with a latent query, while 3D convolution step itself is slower than our self-attention even with  Minkowski Engine. Moreover, 3D CNN requires non-bachify CPU-based voxel ID assignment for each Gaussian primitive, making 3D CNN even slower. We also evaluated our method on text-to-(3D scene) with FID metric, which is 28.32 calculated on rendered views randomly sampled over the unit sphere around 3D scenes, while PointTransformer achieves 153.76.  Although we showcase 3DGS generation for general scenes as an application, the main focus of our paper is about 3D tokenization and latent modeling of scene-level 3D Gaussians. Accordingly, our emphasis is directed towards latent analysis similar to image-to-latent analysis~\cite{van2008visualizing} and the visualization of VAE reconstructions but on unseen inputs, similar to experimental results in Fig.4 of PerceiverIO~\cite{jaegle2021perceiverio}.

\begin{figure}[th]
    \centering
    \begin{minipage}[t]{0.9\linewidth} 
        \centering
        \includegraphics[width=\linewidth]{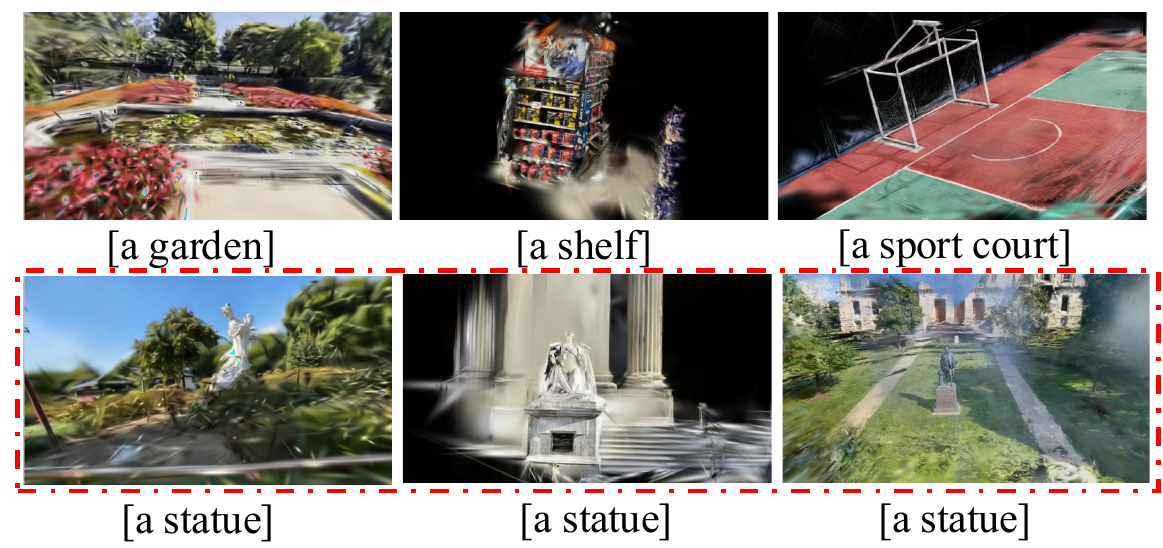} 
    \end{minipage}
    \vspace{-2mm}
    \caption{
        More generative results of our method with corresponding text conditions. Each prompt is intentionally brief and was not seen in the exact form during training to avoid bias toward any specific scene. The results highlight both inter-class and intra-class diversity, with the latter emphasized in red rectangles.
    }
    \label{fig:diversity}
\end{figure}

\newpage
\clearpage
\begin{figure}[ht]
    \centering
    \begin{minipage}[t]{1.0\linewidth} 
        \centering
        \includegraphics[width=\linewidth]{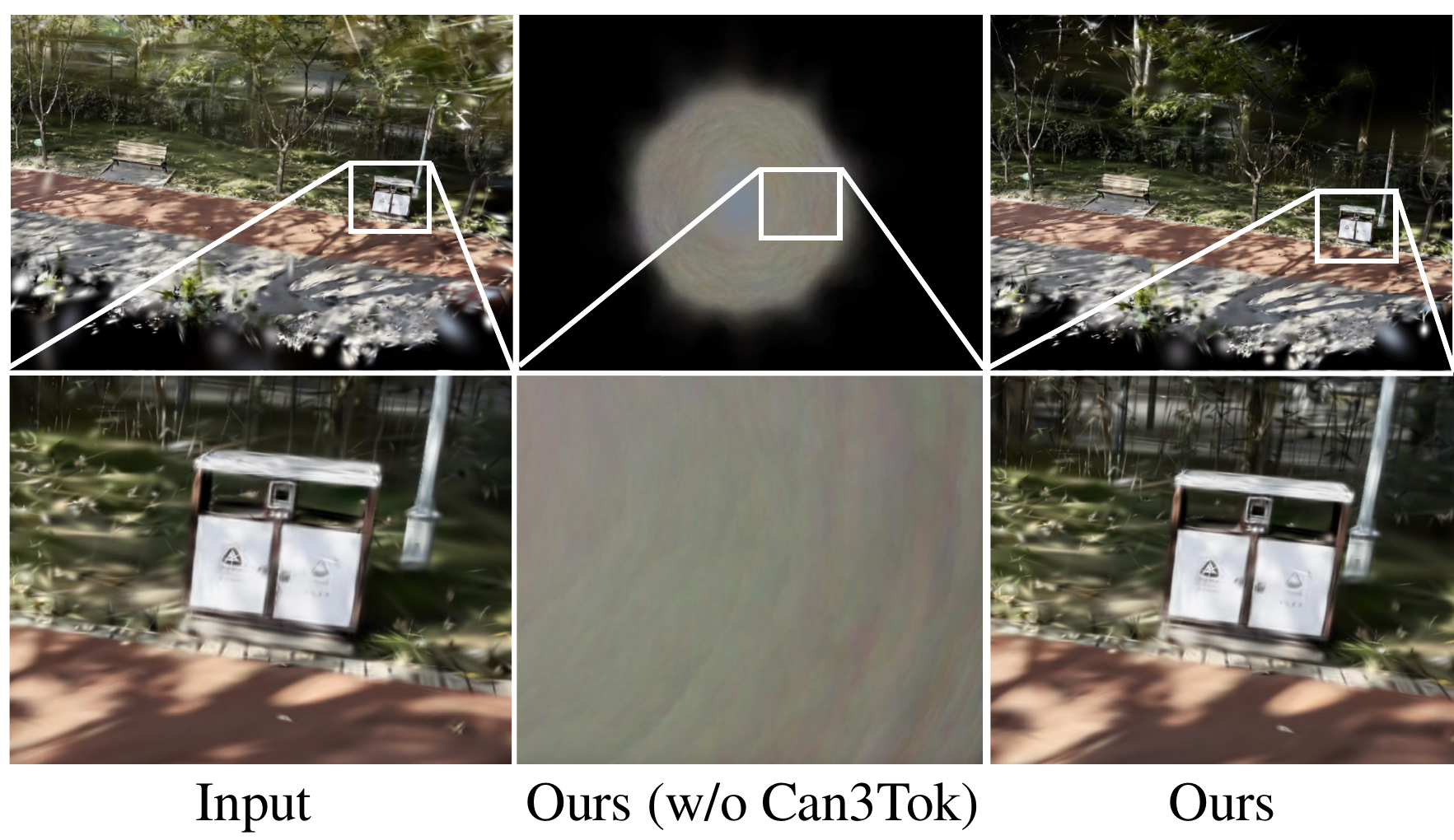} \\ 
    \end{minipage}

    \begin{minipage}[t]{1.0\linewidth} 
        \centering
        \includegraphics[width=\linewidth]{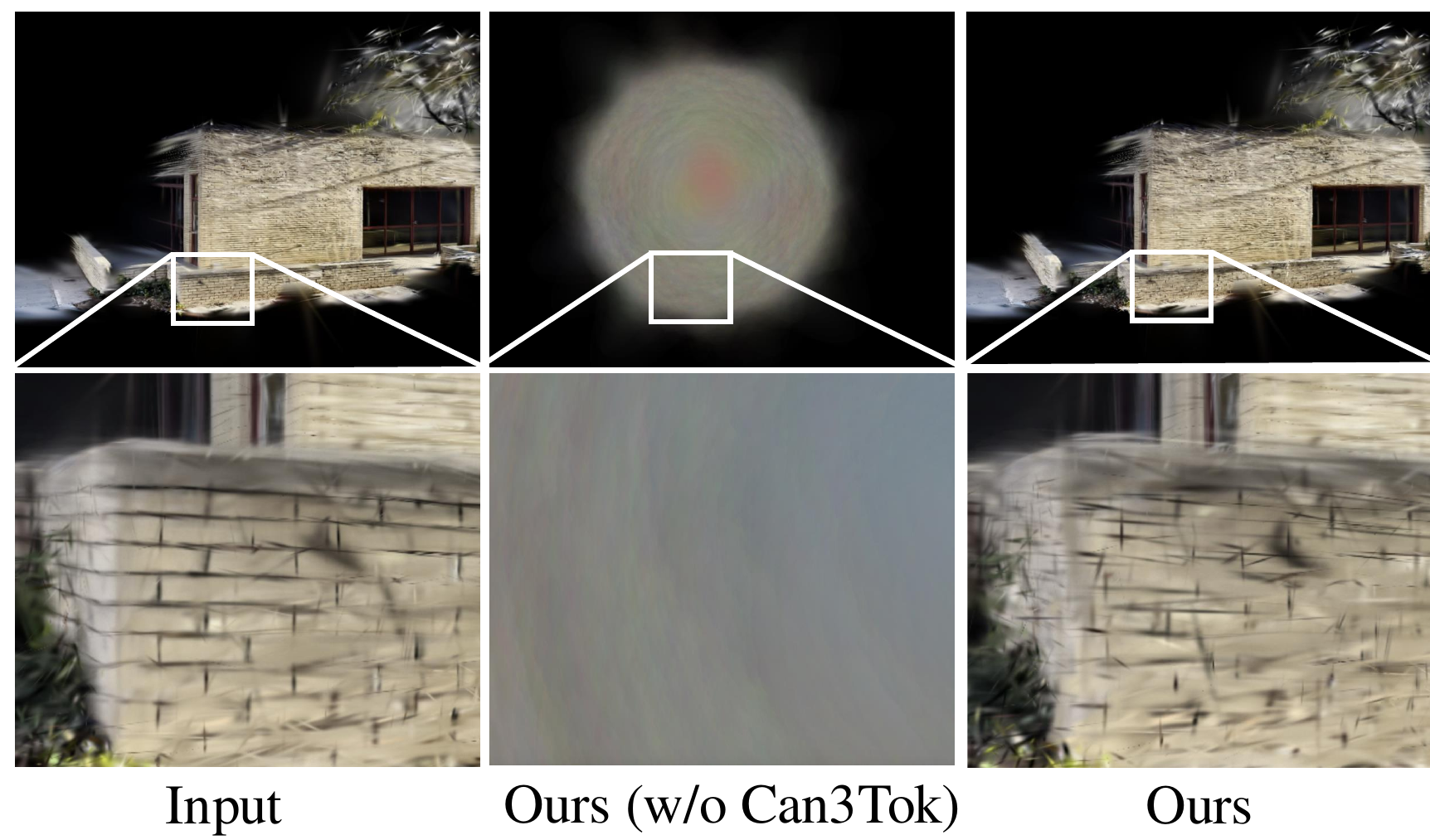} \\ 
    \end{minipage}

    \caption{Qualitative comparisons of w/ and w/o Can3Tok.
    }
    \label{fig:can3tok_ablation}
\end{figure}
\begin{figure}[ht]
    \centering
    \begin{minipage}[t]{1.0\linewidth} 
        \centering
        \includegraphics[width=\linewidth]{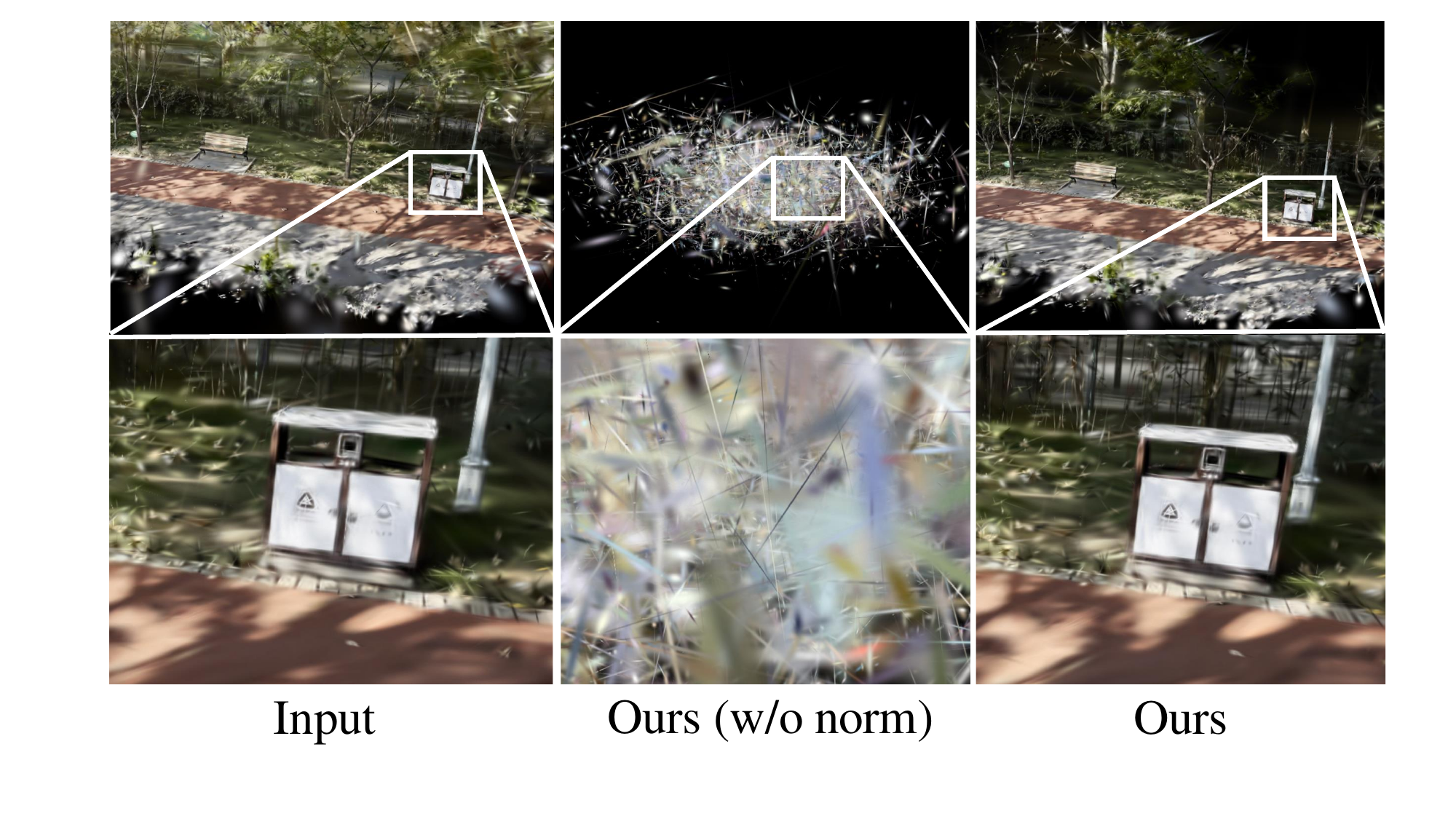} \\ 
    \end{minipage}

    \begin{minipage}[t]{1.0\linewidth} 
        \centering
        \includegraphics[width=\linewidth]{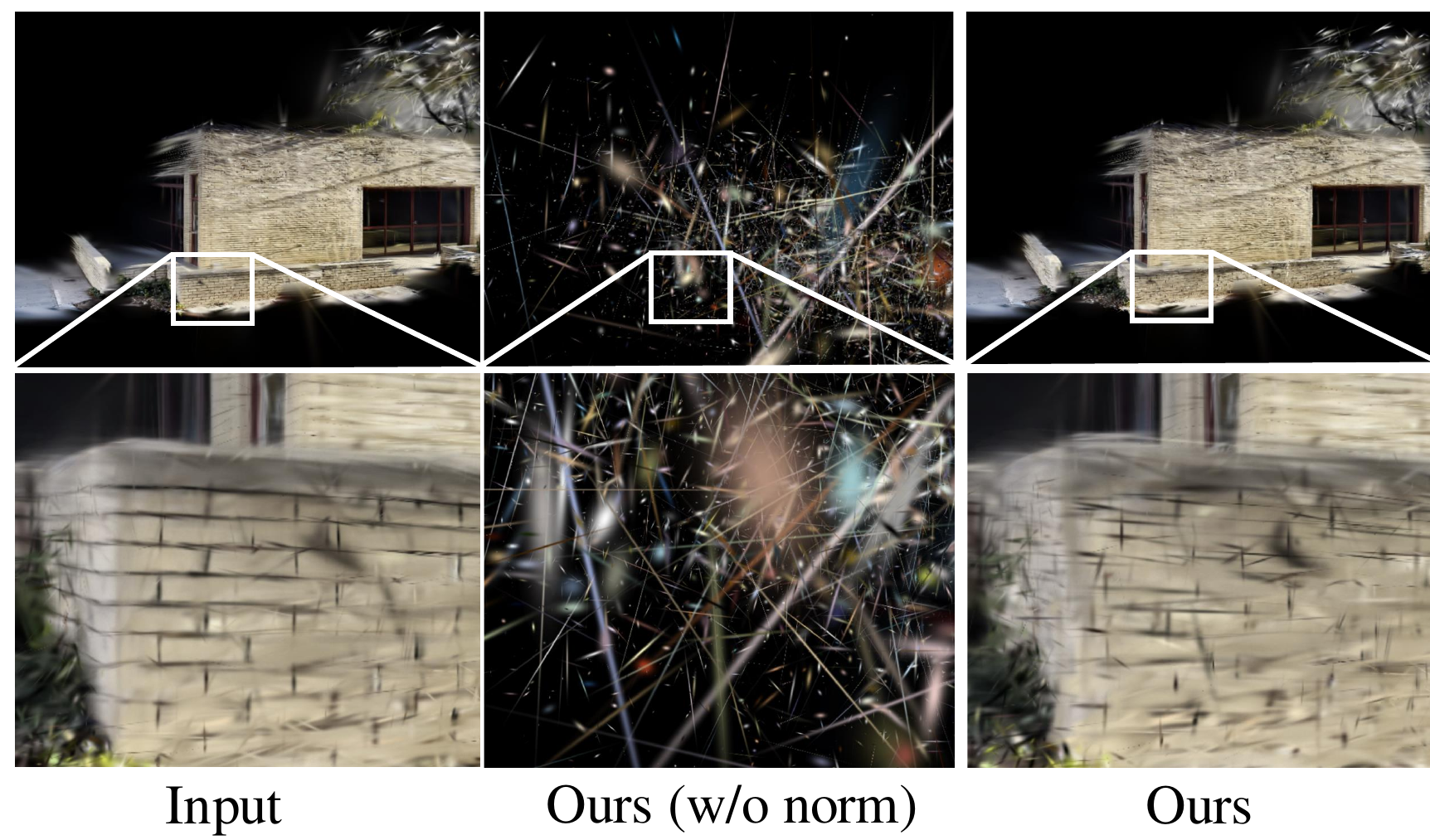} \\ 
    \end{minipage}
    \caption{Qualitative comparisons w/ and w/o normalization.
    }
    \label{fig:norm_ablation}
\end{figure}

\begin{figure}[ht]
    \centering
    \begin{minipage}[t]{1.0\linewidth} 
        \centering
        \includegraphics[width=\linewidth]{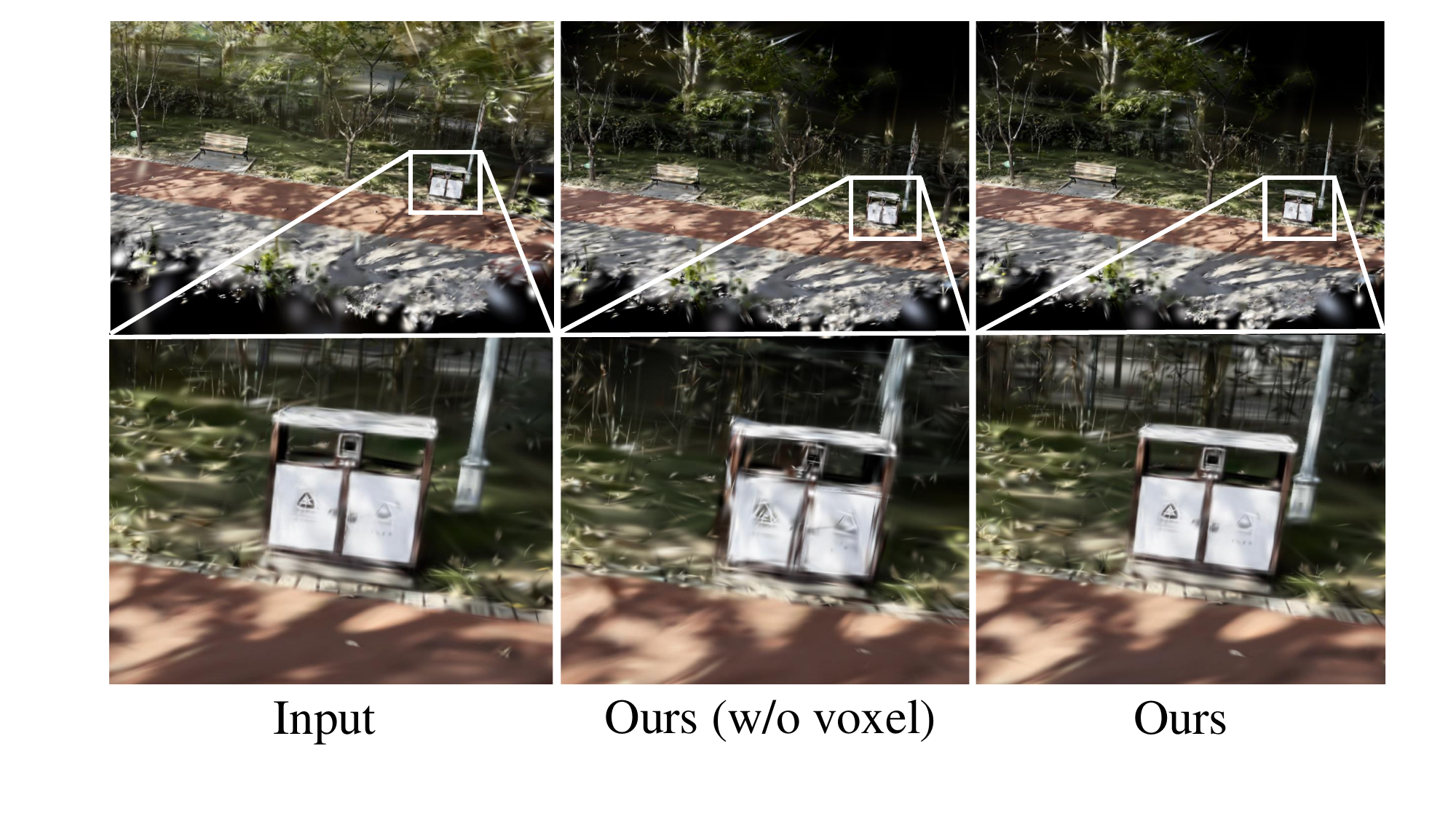} \\ 
    \end{minipage}

     \begin{minipage}[t]{1.0\linewidth} 
        \centering
        \includegraphics[width=\linewidth]{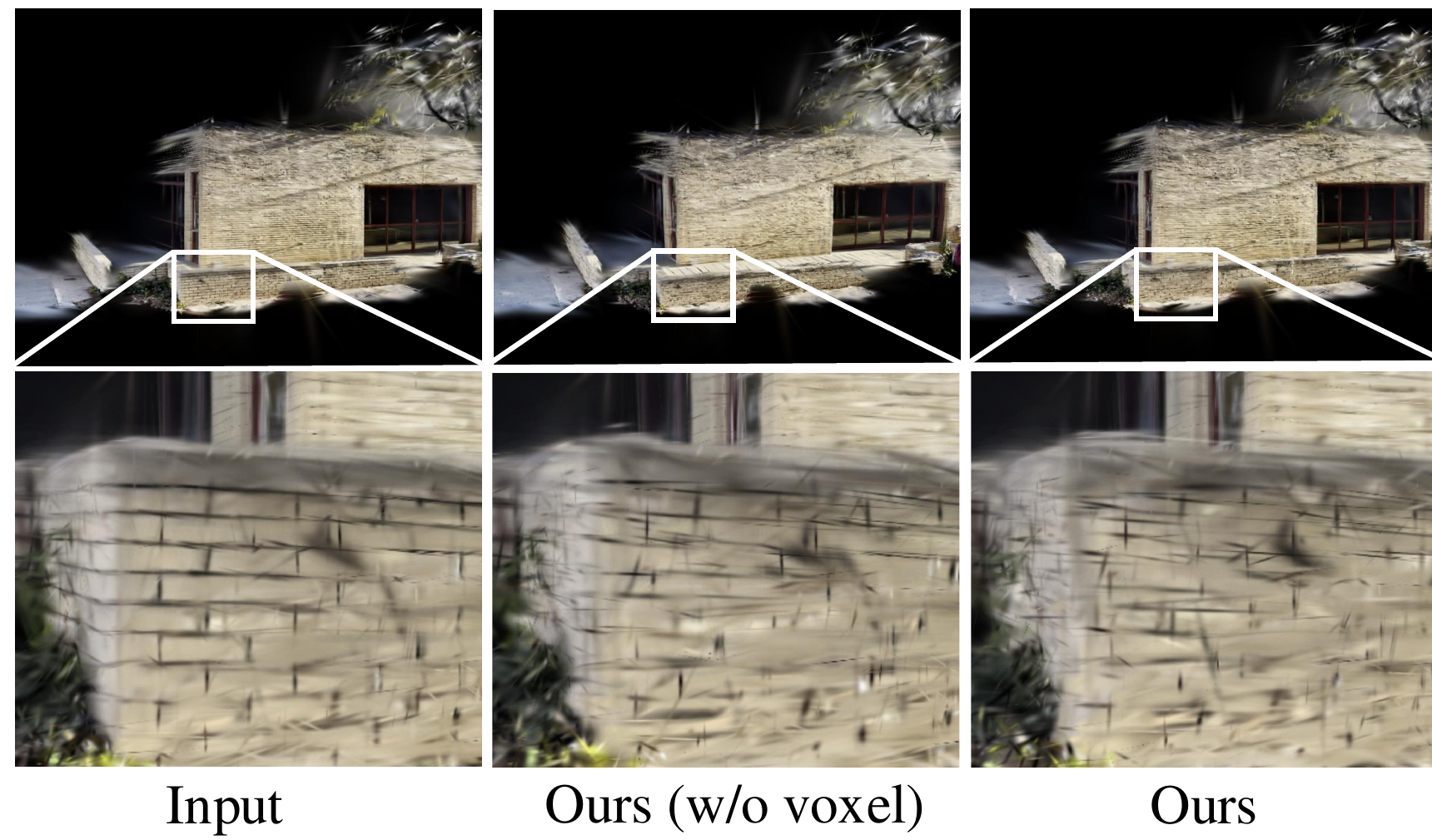} \\ 
    \end{minipage}
    \caption{Qualitative comparisons of w/ and w/o voxel coordinate appending.
    }
    \label{fig:voxel_ablation}
\end{figure}

\begin{figure}[ht]
    \centering
    \begin{minipage}[t]{1.0\linewidth} 
        \centering
        \includegraphics[width=\linewidth]{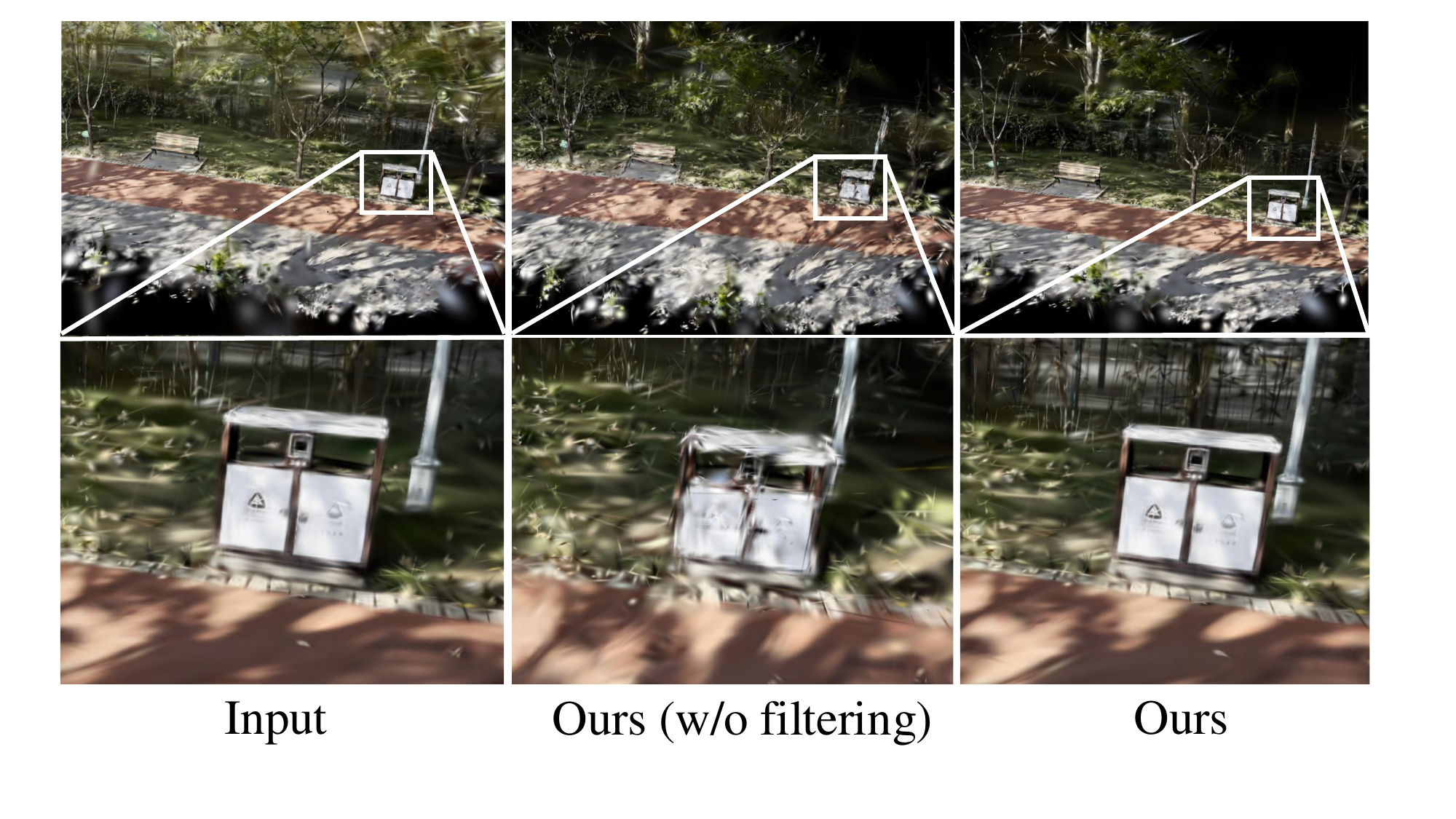} \\ 
    \end{minipage}

    \begin{minipage}[t]{1.0\linewidth} 
        \centering
        \includegraphics[width=\linewidth]{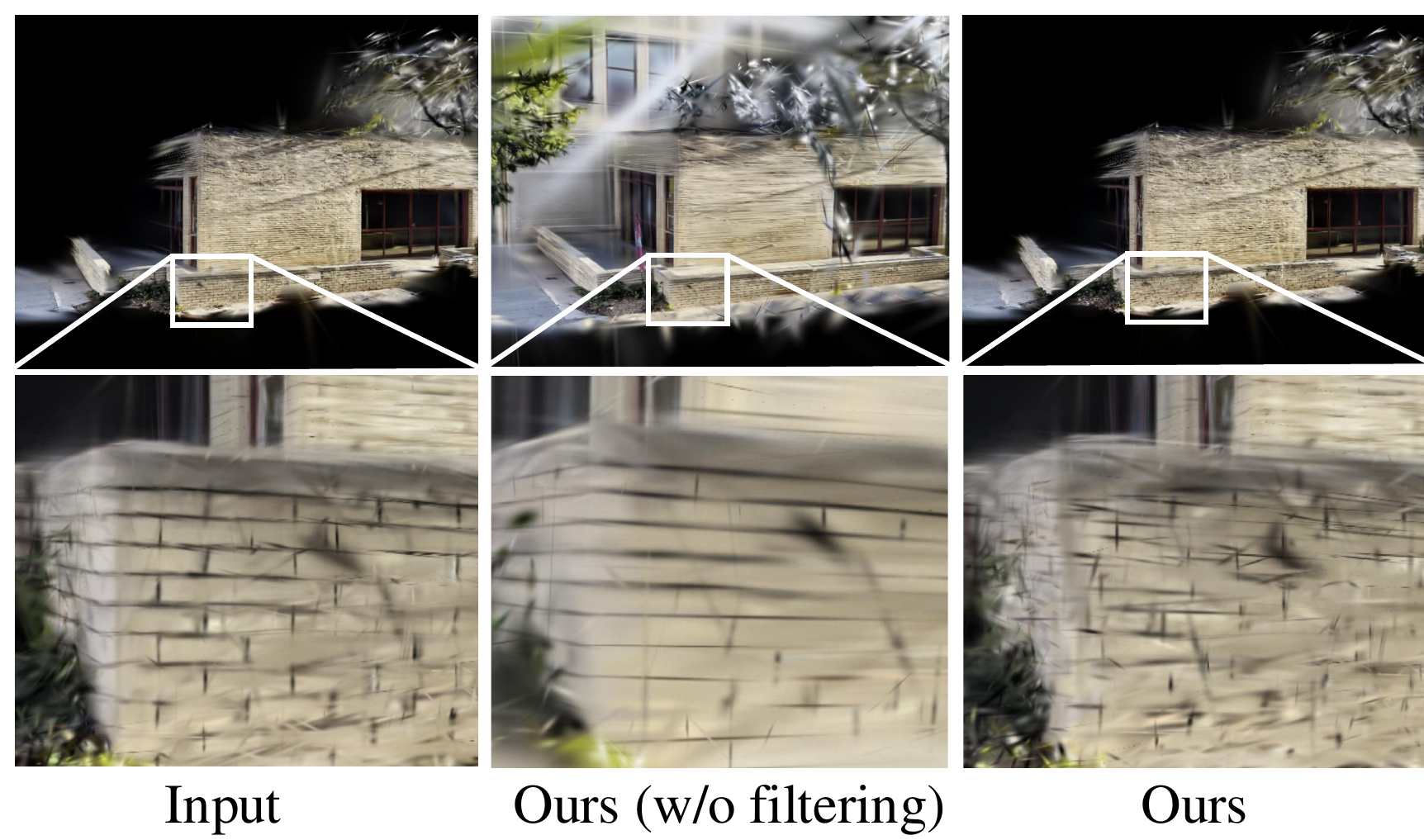} \\ 
    \end{minipage}
    \caption{Qualitative comparisons of w/o semantic-aware 3DGS filtering.
    }
    \label{fig:filtering_ablation}
\end{figure}

\begin{figure*}[ht]
    \centering
    \begin{minipage}[t]{1\linewidth} 
        \centering
        \includegraphics[width=\linewidth]{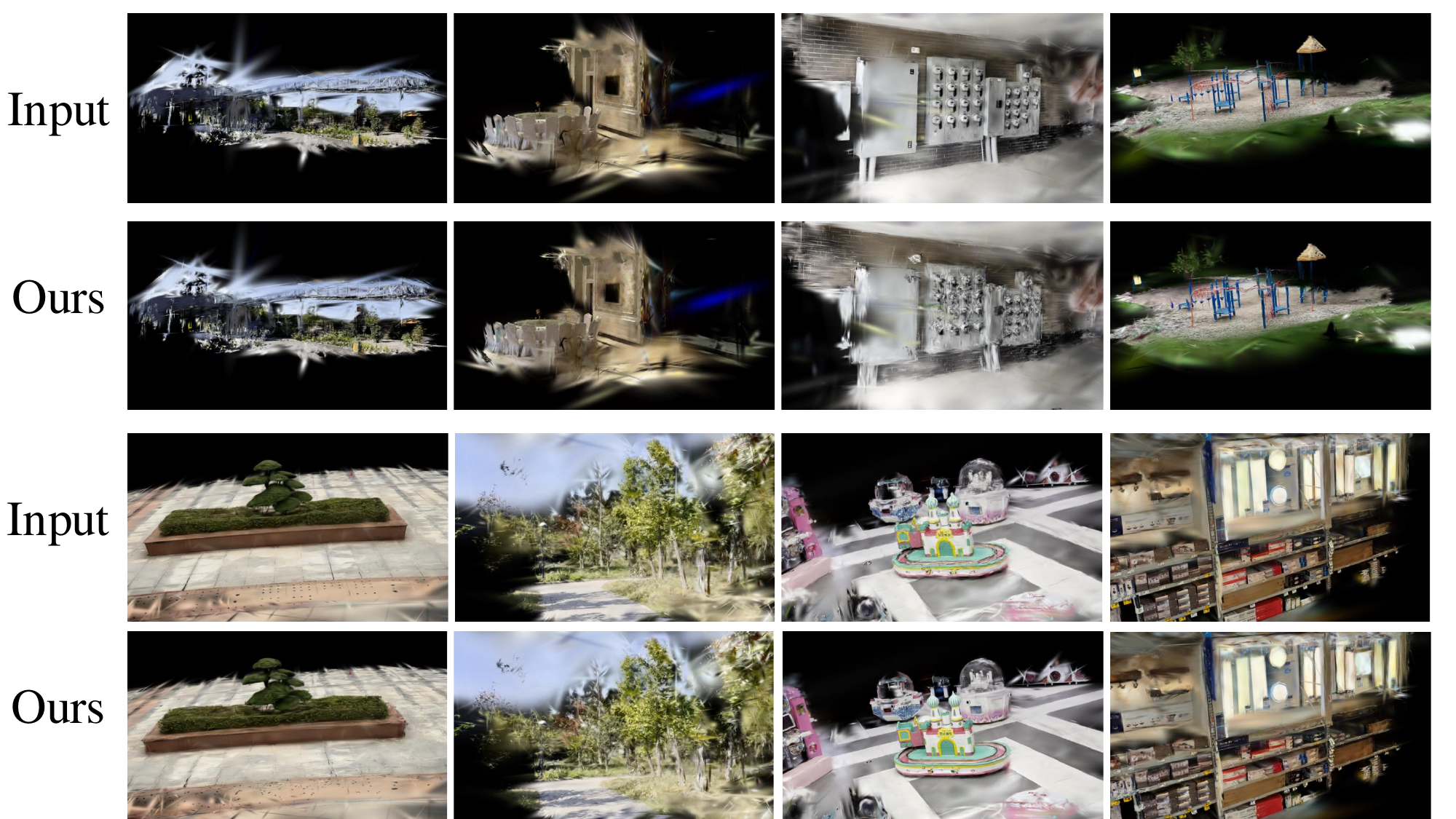} \\ 
    \end{minipage}
    \begin{minipage}[t]{1.0\linewidth} 
        \centering
        \includegraphics[width=\linewidth]{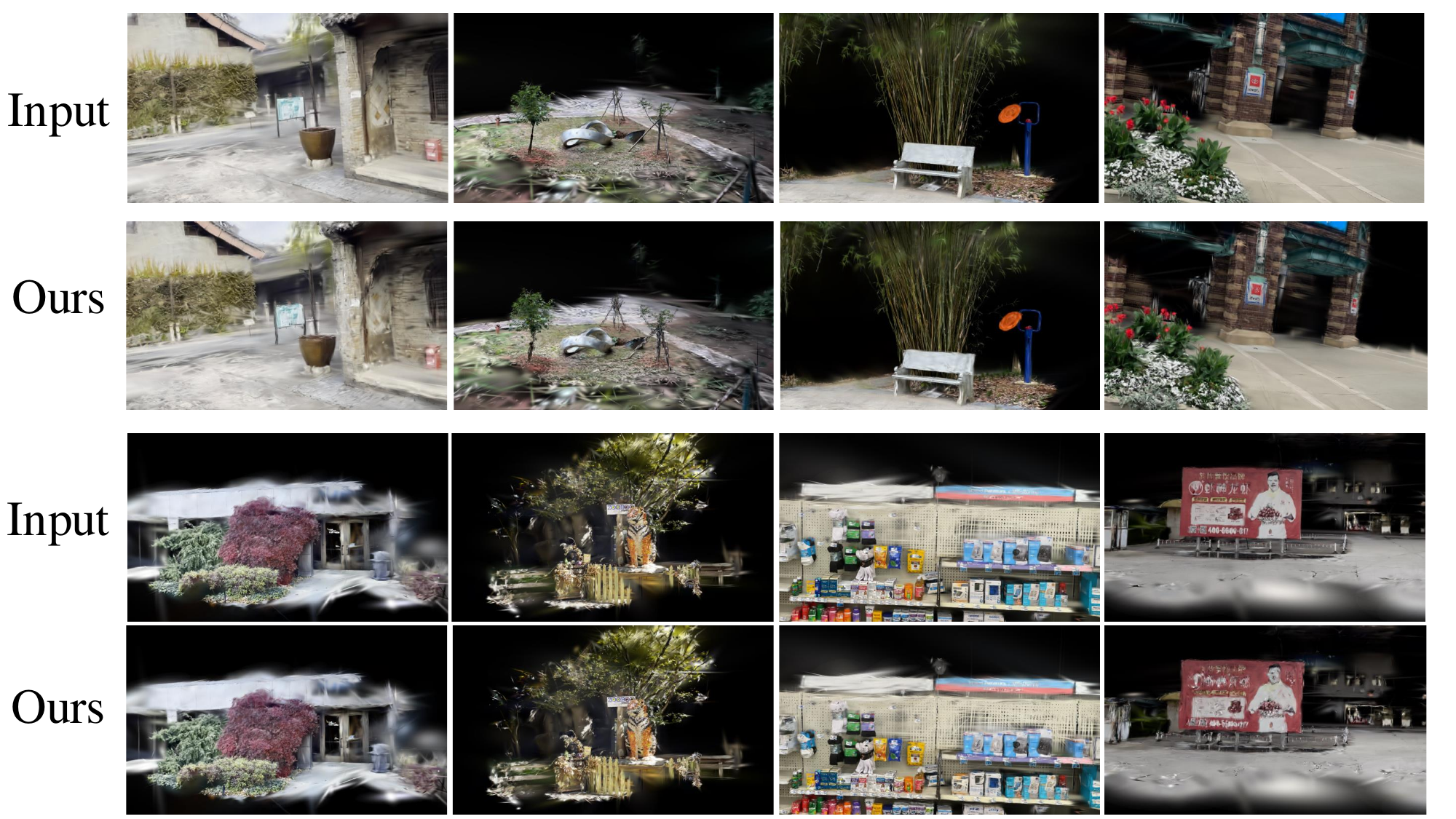} \\ 
    \end{minipage}
    \caption{More qualitative results.
    }
    \label{fig:more_quali}
\end{figure*}

\end{document}